\documentclass[a4paper,fleqn]{cas-dc}
\let\printorcid\relax 
\usepackage[numbers,sort&compress]{natbib}  
\usepackage[ruled,vlined,linesnumbered]{algorithm2e}
\usepackage{placeins}
\usepackage{float} 
\usepackage{amssymb}
\usepackage{dcolumn}

\def\tsc#1{\csdef{#1}{\textsc{\lowercase{#1}}\xspace}}
\tsc{WGM}
\tsc{QE}
\tsc{EP}
\tsc{PMS}
\tsc{BEC}
\tsc{DE}

\begin{document}
\let\WriteBookmarks\relax
\def\floatpagepagefraction{1}
\def\textpagefraction{.001}

\shorttitle{Intelligent Communication Mixture-of-Experts Boosted-Medical Image Segmentation Foundation Model}

\shortauthors{Xinwei Zhang et~al.}

\title [mode = title]{Intelligent Communication Mixture-of-Experts Boosted-Medical Image Segmentation Foundation Model}   

\author[]{Xinwei Zhang\textsuperscript{a}}
\author[]{Hu Chen\textsuperscript{b,*}}
\author[]{Zhe Yuan\textsuperscript{a}}
\author[]{Sukun Tian\textsuperscript{b}}
\author[]{Peng Feng\textsuperscript{a,*}}

\address[a]{College of Optoelectronics Engineering, Chongqing University, Chongqing, China}
\address[b]{Center of Digital Dentistry, Peking University School and Hospital of Stomatology \&{} NHC Key Laboratory of Digital Stomatology, Beijing, China}

\cortext[cor1]{Corresponding authors.}
\fntext[fn1]{E-mail addresses: Chenhu44@126.com (Hu Chen), coe-fp@cqu.edu.cn (Peng Feng)}

\begin{abstract}
Foundation models for medical image segmentation have achieved remarkable performance. Adaptive fine-tuning of natural image segmentation foundation models is crucial for medical image segmentation tasks. However, some limitations exist in existing fine-tuning methods: 1) insufficient representation of high-level features and 2) the fine-tuning process disrupts the structural integrity of pretrained weights. Inspired by these critical problems, we propose an intelligent communication mixture-of-experts boosted-medical image segmentation foundation model, named IC-MoE, with twofold ideas: 1) We construct basic experts, semantic experts, and adaptive experts. Moreover, we implement a pixel probability adaptive voting strategy, which enables expert selection and fusion through label consistency and load balancing. This approach preliminarily enhances the representation capability of high-level features while preserving the structural integrity of pretrained weights. 2) We propose a semantic-guided contrastive learning method to address the issue of weak supervision in contrastive learning. This method further enhances the representation capability of high-level features while preserving the structural integrity of pretrained weights. Extensive experiments across three public medical image segmentation datasets demonstrate that the IC-MoE outperforms other SOTA models. Consequently, the proposed IC-MoE effectively supplements foundational medical image segmentation models with high-level features and pretrained structural integrity. We also validated the superior generalizability of the IC-MoE across diverse medical image segmentation scenarios.
\end{abstract}




\maketitle

\section{Introduction}

Medical image segmentation plays a crucial role in disease screening, clinical diagnosis, and lesion monitoring \cite{de2018clinically,ouyang2020video}. With the advancement of deep learning, researchers have proposed numerous task-specific segmentation models \cite{ronneberger2015u,cao2022swin,isensee2021nnu,chen2021transunettransformersmakestrong}. However, these models rely heavily on large-scale, high-quality labeling data and exhibit limited generalizability. In recent years, the emergence of vision foundation models has driven the development of general-purpose segmentation techniques, such as the Segment Anything Model (SAM) \cite{caron2021emerging,radford2021learning,kirillov2023segment,wang2023seggpt}. Nevertheless, when SAM is directly applied to medical imaging tasks, it often does not perform satisfactorily \cite{he2023accuracy}. Consequently, researchers have attempted knowledge transfer through parameter-efficient fine-tuning methods \cite{gu2024build}. However, current fine-tuning approaches have two central problems: 1) parameter-efficient fine-tuning methods lack the ability to capture high-level features of complex medical structures, and 2) parameter-efficient fine-tuning methods disrupt the structural integrity of pretrained weights.

\begin{figure*}[t]
    \centering
    \includegraphics[width=\linewidth,keepaspectratio]{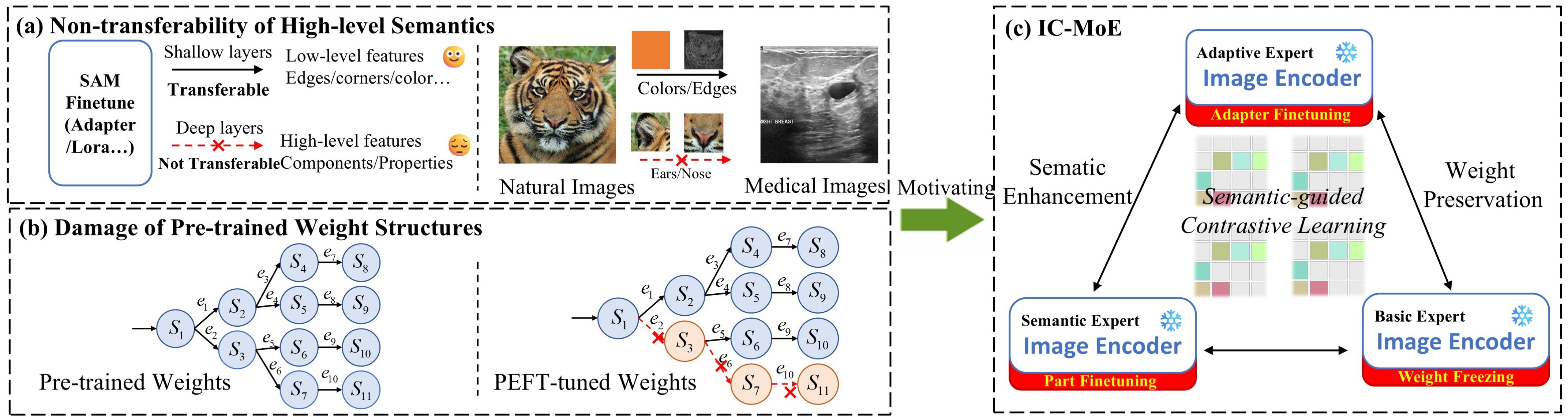}
    \caption{The Motivation of the Proposed IC-MoE. (a) Common fine-tuning methods fail to transfer high-level semantic representation capabilities. (b) Common fine-tuning methods disrupt the structural integrity of pretrained weights. (c) The IC-MoE method both enhances high-level features and preserves the structural integrity of pretrained weights.}
    \label{FIG:1}
\end{figure*}

First, parameter-efficient fine-tuning strategies fail to address the issue of inadequate high-level features in medical images. While fine-tuning effectively transfers low-level features learned in shallow layers, it struggles to transfer high-level features learned in deep layers. The high-level semantic knowledge learned from natural images—such as “nose” and “ears”—cannot be directly applied to recognize complex organs such as “tumors”. See Figure~\ref{FIG:1}(a). Zhang et al.\cite{zhang2023customized} employed the LoRA fine-tuning method for weight fine-tuning; Adapter fine-tuning methods have been widely used for domain transfer \cite{wu2025medical,cheng2023sammed2d,feng2025swinsam}; Chen et al.\cite{chen2023sam} combined adapter fine-tuning with frequency domain information to supplement domain-specific knowledge ; Zhong et al.\cite{zhong2024convolution} fused convolutional structures with LoRA methods to enhance feature extraction in local spaces; Wei et al.\cite{wei2024stronger} refined interlayer features through trainable parameters . While these approaches mitigate the semantic gap in different ways, they remain insufficient for learning the high-level features required in medical images.

Second, parameter-efficient fine-tuning methods disrupt the structural integrity of pretrained weights. In the original model architecture, each layer's input originates from the preceding layer. The introduction of fine-tuning modules leads to poor utilization of weights in certain branch structures. See Figure~\ref{FIG:1}(b). In addition to normal fine-tuning approaches, methods that introduce additional branches to supplement high-level semantics also have this problem. For example, Lin et al.\cite{lin2023samus} and Lin et al.\cite{lin2024samct} designed CNN branches outside the SAM architecture to increase the encoder's segmentation accuracy for minute lesions and blurred boundaries; Liang et al.\cite{liang2025mambasam} designed the VMamba branch to strengthen the model's ability to capture medical image details. While these approaches supplement high-level features, they all destroy the structural integrity of pretrained weights. Besides, Wang et al.\cite{wang2024sam} introduced two branches—domain-specific experts and original experts—in the decoder, yet these experts lack information communication. This approach fails to fully integrate general knowledge with domain-specific knowledge, resulting in limited performance improvement. Zhang et al. \cite{zhang2024improving} incorporated a pretrained weight structure into a self-training framework via weakly supervised contrastive loss. However, insufficient weak supervision signals constrain the effectiveness of feature regularization. In summary, existing approaches fail to address both the insufficiency of high-level features and the disruption of pretrained weight structures simultaneously. Furthermore, the lack of effective information exchange in weakly supervised contrastive losses limits the full utilization of high-level semantic features. These issues highlight the current limitations of fine-tuning methods.

To address the above challenges, we propose the IC-MoE foundational model. This model comprises three complementary expert modules: the basic expert, the semantic expert, and the adaptive expert. Simultaneously, we design an adaptive voting strategy based on pixel-level probabilities to dynamically fuse outputs from these three experts. This framework effectively preserves the structural integrity of pretrained weights while augmenting high-level features. Furthermore, we introduced a semantic-guided contrastive learning approach to address the problem of insufficient supervisory information in contrastive learning. Through these methods, the IC-MOE model successfully enhances high-level semantic information while retaining the structural integrity of pretrained weights.

The main contributions of this paper are as follows:
\begin{enumerate}
    \item We construct the basic expert, the semantic expert, and the adaptive expert, along with a pixel probability adaptive voting strategy. We implement an intelligent communication mixture-of-experts approach to mitigate the insufficient representation of high-level features and preserve the structural integrity of pretrained weights.
    \item We propose a semantic-guided contrastive learning approach to address the weak supervision issue in contrastive learning. This method enables intelligent communication among expert models, enhancing high-level semantic information while preserving the structural integrity of pretrained weights.
\end{enumerate}

\section{METHODOLOGY}

\subsection{Overview of the Intelligent Communication Mixture-of-Experts Boosted-Medical Image Segmentation Foundation Model}
Current fine-tuning methods often fail to capture sufficient high-level semantic information and tend to compromise the structural integrity of pretrained weights. To address this, we propose the intelligent communication mixture-of-experts boosted-medical image segmentation foundation model (IC-MoE, See Figure~\ref{FIG:2}). 
First, we divide the input image $\mathbf{X}_{i}$ into three parts: $\mathbf{X}_{i}=[\mathbf{X}_{i}^{img},\mathbf{X}_{i}^{fg},\mathbf{X}_{i}^{bg}]$. 
Second, we feed $\mathbf{X}_{i}$ into the ECFM module to obtain three sets of expert feature tensors: $\mathbf{Y}_{e}=[\mathbf{Y}_{e}^{img},\mathbf{Y}_{e}^{fg},\mathbf{Y}_{e}^{bg}]$. To reduce computational costs, we disable gradient propagation during the foreground and background feature extraction process. We then fuse these features to produce the average feature $\mathbf{Y}_{3}=[\mathbf{Y}_{3}^{img},\mathbf{Y}_{3}^{fg},\mathbf{Y}_{3}^{bg}]$. Third, features from all the experts are input into the SgCL module for semantic-guided contrastive learning. This module significantly enhances the representation of high-level semantics while preserving the structural integrity of pretrained weights. Fourth, we feed the expert features into the decoder for prediction and obtain the probability map $\mathbf{P}_{e}=[\mathbf{P}_{0},\mathbf{P}_{1},\mathbf{P}_{2}]$. Here, we set $\mathbf{P}_{2}$ as the main expert and the others as candidate experts. Fifth, in the PPAV module, we select the optimal candidate expert for each sample on the basis of pixel probabilities. Finally, the segmentation results from the candidate expert and the main expert are fused to obtain the final segmentation mask.

\subsection{Expert Communication and Fusion Module}
\begin{figure*}[t]
    \centering
    \includegraphics[width=\linewidth,keepaspectratio]{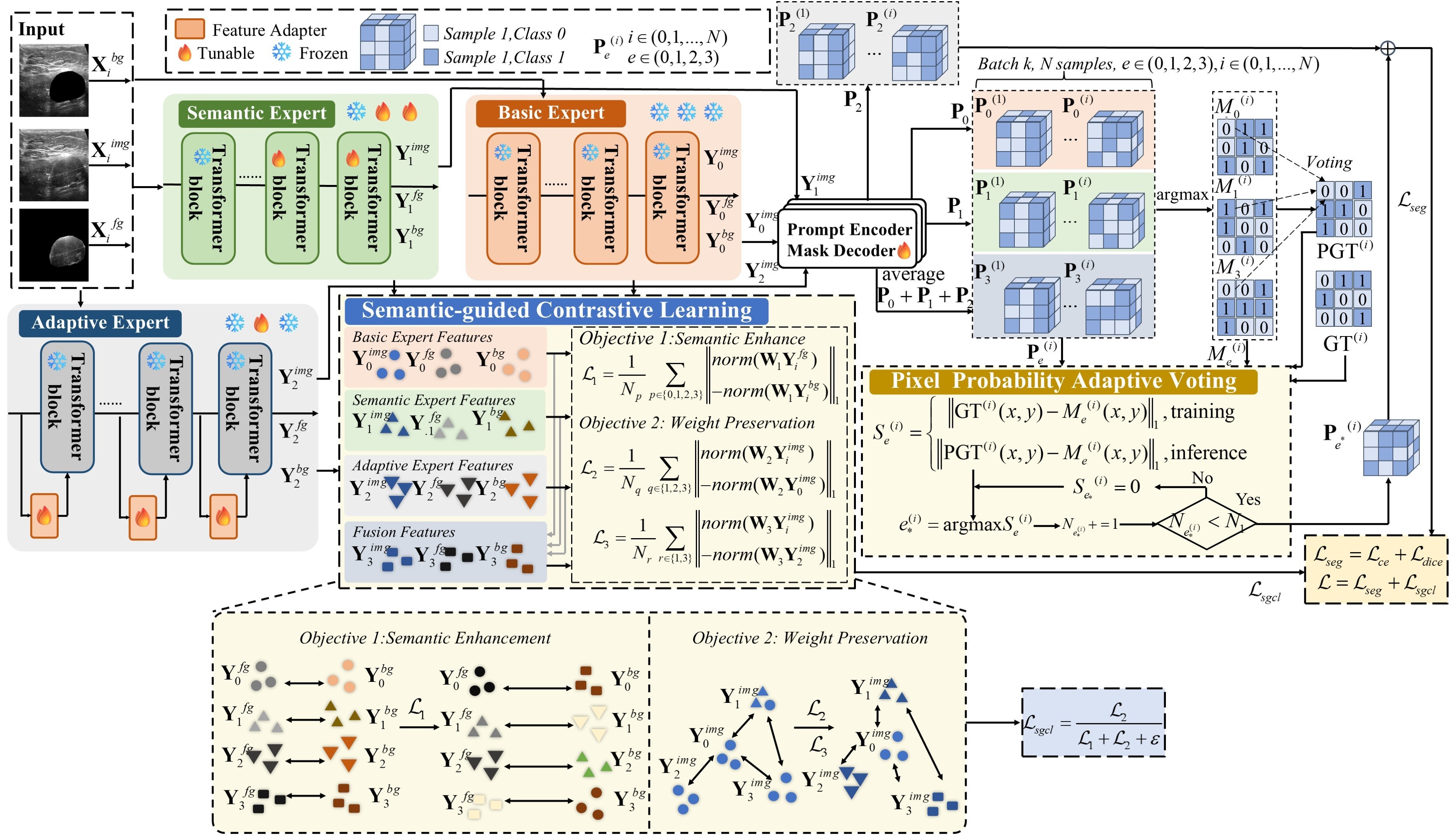}
    \caption{Overview of the proposed IC-MoE foundation model.}
    \label{FIG:2}
\end{figure*}

According to the Introduction, our task is to address the problem of insufficient high-level features in fine-tuning methods. Additionally, we aim to maintain the structural integrity of pretrained weights during fine-tuning. To address both problems, we first propose the Expert Communication and Fusion Module (ECFM). The overall workflow is shown in Figure~\ref{FIG:2} and Algorithm~\ref{alg:ppav}. The ECFM consists of two components: basic-semantic-adaptive experts (BSAEs) and pixel probability adaptive voting (PPAV).

\noindent\textbf{Basic-Semantic-Adaptive Experts.}  We build three functionally complementary experts within the BSAEs, each retaining the same architecture as the original SAM: the image encoder, prompt encoder, and mask decoder. The basic expert retains the original pretrained weights of the SAM model without fine-tuning. It leverages the SAM's powerful feature extraction capability to maintain the structural integrity of pretrained weights within the model. Semantic experts focus on learning complex high-level features in medical images. We fine-tune only the last two transformer blocks within the image encoder. This strategy enables the expert to optimize high-level features relevant to medical images while preserving low-level general feature extraction capabilities. Adaptive Expert serves as the model backbone and is designed to efficiently transfer general knowledge of specific medical image tasks. By inserting lightweight adapter modules into each transformer block, this expert efficiently adapts to medical image tasks. The introduction of the BSAEs ensures the structural integrity of pretrained weights during adaptation.

\begin{algorithm}[t]  
\caption{Pixel Probability Adaptive Voting}
\label{alg:ppav}
\KwIn{Predictions $\{\mathbf{P}_0, \mathbf{P}_1, \mathbf{P}_2\}$, Ground truth $GT$}
\KwOut{Final logits $\mathbf{P}_{final}^{(i)}$}

$\mathbf{P} \leftarrow \frac{1}{3}\sum\nolimits_{e=0}^{2} \mathbf{P}_e$,\quad
$\mathbf{P}_{base} \leftarrow \mathbf{P}_2$,\quad
$\mathcal{C} \leftarrow \{\mathbf{P}_0, \mathbf{P}_1, \mathbf{P}_3\}$\;

Initialize counts: $N_e \leftarrow 0$, for $e \in \{1,2,3\}$\;

\For{each sample $i$ in the batch}{
    Obtain pseudo ground truth $\text{PGT}^{(i)}$ 
    by candidate expert voting (see Eq.~\ref{Eq:2})\;
    
    Compute each candidate score $S_e^{(i)}$ 
    for $e \in \mathcal{C}$ (see Eq.~\ref{Eq:3})\;
    
    $e_*^{(i)} \leftarrow \arg\max S_e^{(i)}$\;
    \While{$N_{e_*^{(i)}} > N_\text{threshold}$}{
        $S_{e_*}^{(i)} \leftarrow 0$\;
        $e_*^{(i)} \leftarrow \arg\max S_e^{(i)}$\;
    }
    $N_{e_*^{(i)}} \leftarrow N_{e_*^{(i)}} + 1$\;
    
    $\mathbf{P}_{final}^{(i)} \leftarrow 
    \mathbf{P}_{base}^{(i)} + \alpha \times \mathbf{P}_{e_*^{(i)}}^{(i)}$\;
}
\end{algorithm}

\begin{figure*}[t]
    \centering
    \includegraphics[width=\linewidth,keepaspectratio]{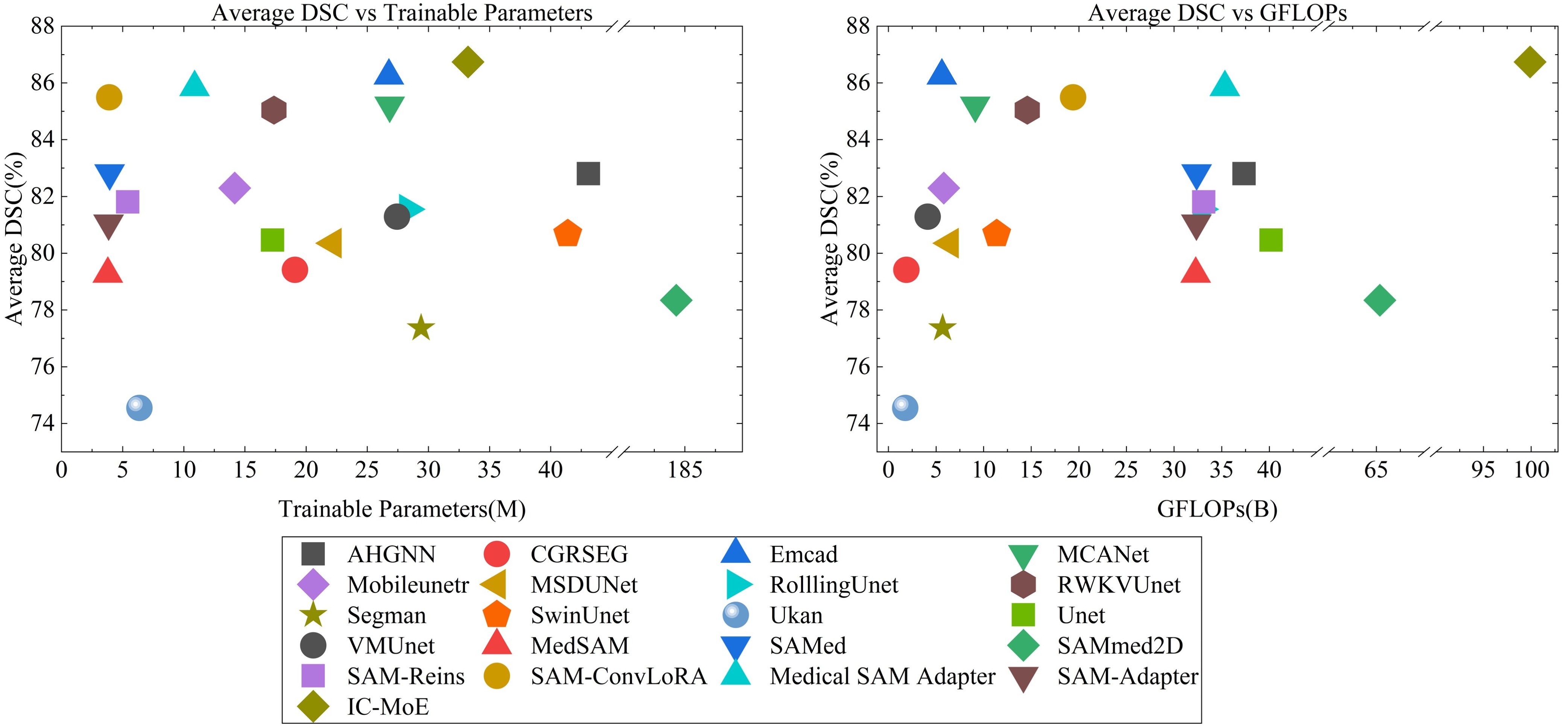}
    \caption{Average DSC vs. trainable parameters and average DSC vs. GFLOPs.}
    \label{FIG:3}
\end{figure*}

\noindent\textbf{Pixel probability adaptive voting.} First, we introduce the fusion result $\mathbf{P}_{3}$ to enhance stability. We set $\mathbf{P}_{2}$ as the primary expert and the other experts as candidate experts:
\begin{equation}
\mathcal{C}=\{{\mathbf{P}_{0},\mathbf{P}_{1},\mathbf{P}_{3}}\}
\end{equation}
Second, we used an adaptive voting strategy to select the optimal expert for each sample. In the training process, we use the ground truth as the supervision signal. In the inference process, we use the pseudo ground truth PGT as the supervision signal. The label for each pixel in PGT is voted on by the candidate experts; see Eq.~(\ref{Eq:2}). Then, we score each candidate expert by calculating the similarity between prediction masks 
$M_{e}^{(i)}(x,y)$ and $\text{GT}^{(i)}$/$\text{PGT}^{(i)}$; see Eq.~(\ref{Eq:3}).

\begin{equation}
	\text{PGT}(x,y)=
    \begin{cases}
1, & \text{if } \sum_{e \in \{0,1,3\}} M_e^{(i)}(x,y) > 1\\
0, & \text{otherwise}
\end{cases}
\label{Eq:2}
\end{equation}

where $M_{e}^{(i)},e\in (0,1,3),i\in (0,1,2,...,N)$ denotes expert e's prediction mask for sample i.

\begin{equation}
	S_{e}^{(i)}=
    \begin{cases}
   {{\left\| \text{G}{{\text{T}}^{(i)}}(x,y)-M_{e}^{(i)}(x,y) \right\|}_{1}},\text{training}  \\
   {{\left\| \text{PG}{{\text{T}}^{(i)}}(x,y)-M_{e}^{(i)}(x,y) \right\|}_{1}},\text{inference} 
\end{cases}
\label{Eq:3}
\end{equation}

where $\text{GT}^{(i)}$ denotes the ground truth mask for sample i and where $\text{PGT}^{(i)}$ denotes the prediction mask for sample i.

Third, to ensure training stability, we introduce a balancing strategy. This strategy guarantees that each candidate expert has an equal chance of being selected. First, on the basis of the scores from Step 2, we index the top-scoring expert $e_{*}^{(i)}$; see Eq.~(\ref{Eq:4}). Then, we count the total number of times this expert has been selected. If the selection count exceeds a threshold, we reset this expert's score from the current sample to zero and perform a new selection. Otherwise, we consider this expert the optimal candidate. This mechanism guarantees each expert an equal chance of being selected.

\begin{equation}
    e_{*}^{(i)}=\arg \max S_{e}^{(i)}
\label{Eq:4}
\end{equation}

Fourth, we fuse the prediction result $\mathbf{P}_{{{e}^{*}}}^{(i)}\ $of the best candidate expert with the prediction result $\mathbf{P}_{2}^{(i)}$ of the base expert. We can obtain the segmentation result for the i-th sample:
\begin{equation}
    \mathbf{P}_{final}^{(i)}=\mathbf{P}_{2}^{(i)}+\alpha \times \mathbf{P}_{{{e}^{*}}}^{(i)}
\end{equation}
where $\alpha$ is the fusion coefficient, which represents the contribution of the optimal candidate expert to the base expert.

\subsection{Semantic-guided Contrastive Learning}
According to the Introduction, our task is to address the issue of intelligent communication between expert modules. Additionally, we must address the problem of weak supervisory information in contrastive learning. To this end, we propose semantic-guided contrastive learning, with the overall workflow illustrated in Figure \ref{FIG:1}. Semantic-guided contrastive learning comprises three components: semantic enhancement, weight preservation, and contrastive guidance.

\noindent\textbf{Semantic Enhancement.} The key to medical image segmentation is to accurately separate the foreground from background. To achieve this, we maximize the semantic distance between the foreground and background features. By explicitly computing feature distances, this optimization objective further enhances the model's ability to represent high-level features in medical images. See Eq.~(\ref{Eq:6}).

\begin{equation}
\begin{split}
\mathcal{L}_{1} &= \max\Bigg(
\frac{1}{N_i} \sum_{i \in \{0,1,2,3\}}
\Big\| \mathrm{norm}(\mathbf{W}_1 \mathbf{Y}_i^{fg}) \\
&\quad - \mathrm{norm}(\mathbf{W}_1 \mathbf{Y}_i^{bg}) \Big\|_1
\Bigg)
\end{split}
\label{Eq:6}
\end{equation}

\noindent\textbf{Weight preservation.} Common fine-tuning methods often disrupt the structural integrity of pretrained weights. To address this, we apply two complementary constraints. First, we apply a consistency constraint to minimize the feature distance between other experts and the basic expert. See Eq.~(\ref{Eq:7}).
\begin{equation}
\begin{split}
\mathcal{L}_{2} &= \min \Bigg(
\frac{1}{N_j} \sum_{j \in \{0,1,3\}} 
\Big\| \mathrm{norm}(\mathbf{W}_2 \mathbf{Y}_j^{\mathrm{img}}) \\
&\quad - \mathrm{norm}(\mathbf{W}_2 \mathbf{Y}_2^{\mathrm{img}}) \Big\|_1
\Bigg)
\end{split}
\label{Eq:7}
\end{equation}

To ensure complementarity among experts, we maximize the feature distance between other experts and the adaptive expert. These constraints ensure that experts capture diverse semantic features while preserving the general knowledge embedded in pretrained weights.See Eq.~(\ref{Eq:8}).
\begin{equation}
\begin{split}
\mathcal{L}_{3} &= \max \Bigg(
\frac{1}{N_k} \sum_{k \in \{1,3\}} 
\Big\| \mathrm{norm}(\mathbf{W}_3 \mathbf{Y}_k^{\mathrm{img}}) \\
&\quad - \mathrm{norm}(\mathbf{W}_3 \mathbf{Y}_0^{\mathrm{img}}) \Big\|_1
\Bigg)
\end{split}
\label{Eq:8}
\end{equation}
    
\noindent\textbf{Contrastive guidance.} To achieve these objectives, we design Semantic-guided contrastive learning (SgCL).See Eq.~(\ref{Eq:9}).
\begin{equation}
    {{\mathcal{L}}_{sgcl}}=\frac{{{\mathcal{L}}_{2}}}{{{\mathcal{L}}_{1}}+{{\mathcal{L}}_{3}}+\varepsilon }
\label{Eq:9}
\end{equation}

where $\mathcal{L}_{1}$ enhances high-level features, while 
$\mathcal{L}_{2}$ and $\mathcal{L}_{3}$ supplement the structural integrity of pretrained weights. Semantic-guided contrastive loss significantly enhances communication capabilities among expert modules.
The overall loss function of the model is composed of segmentation losses and , along with semantic-guided contrastive loss :
\begin{equation}
    \mathcal{L}=\alpha \times {{\mathcal{L}}_{ce}}+\beta \times {{\mathcal{L}}_{dice}}+\gamma {{\mathcal{L}}_{sgcl}}
\end{equation}

\begin{figure}
	\centering
		\includegraphics[width=0.5\textwidth]{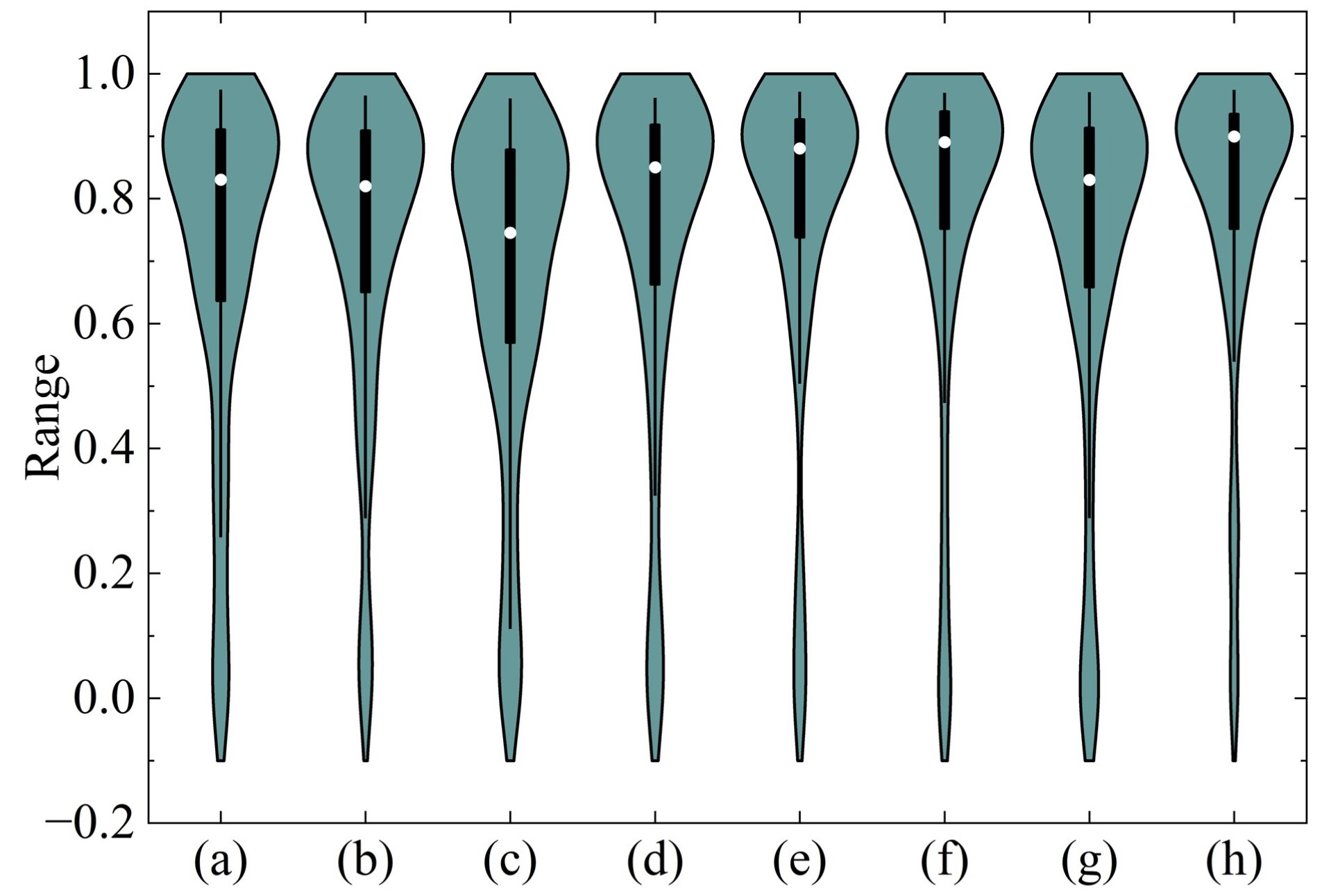}
	\caption{DSC distribution.(a) MedSAM (b) SAMed (c) SAMmed2D (d) SAM-Reins (e) SAM-ConvLoRA (f) Medical SAM Adapter (g) SAM-Adapter (h) IC-MoE}
	\label{FIG:4}
\end{figure}

\begin{table*}[width=\textwidth, cols=8, pos=t]  
\caption{Results of the ablation experiment on the BUSI dataset.}\label{tbl1}
\begin{tabular*}{\tblwidth}{@{} L L L c c c c c @{} }
\toprule
Model / Metrics & ECFM & SgCL & \multicolumn{1}{c}{DSC} & \multicolumn{1}{c}{IoU} & \multicolumn{1}{c}{Accuracy} & \multicolumn{1}{c}{Recall} & \multicolumn{1}{c@{}}{Precision} \\
\midrule
IC-MoE & × & × & 0.7766 & 0.6909 & 0.9582 & 0.8015 & 0.8008 \\
       & $\checkmark$ & × & 0.7786 & 0.6896 & 0.9567 & 0.8173 & 0.7930 \\
       & $\checkmark$ & $\checkmark$ & 0.7926 & 0.7040 & 0.9594 & 0.8299 & 0.7974 \\
\bottomrule
\end{tabular*}
\end{table*}

\begin{table*}[width=\textwidth, pos=t]
\centering
\caption{Segmentation Results of IC-MoE on the ISIC, BUSI, and GLAS Datasets.}\label{tbl2}
\begin{tabular*}{\tblwidth}{@{} C C C C C C C C C @{} }
\toprule
\multirow{2}{*}{\centering Model} & \multicolumn{2}{C}{ISIC 2018} & \multicolumn{2}{C}{BUSI} 
& \multicolumn{2}{C}{GLAS} & \multicolumn{2}{C@{}}{AVG} \\
\cmidrule(lr){2-3} \cmidrule(lr){4-5} \cmidrule(lr){6-7} \cmidrule(l){8-9}
& DSC    & IoU    & DSC    & IoU    & DSC    & IoU    & DSC    & IoU \\
\midrule
AHGNN$_{\textit{MICCAI'24 \cite{chai2024novel}}}$       & 86.99 & 79.17 & 72.25 & 63.02 & 89.17 & 81.16 & 82.80 & 74.45 \\
CGRSEG$_{\textit{ECCV'24 \cite{ni2024context}}}$        & 86.95 & 78.67 & 66.73 & 55.57 & 84.57 & 74.97 & 79.42 & 69.74 \\
Emcad$_{\textit{CVPR'24 \cite{rahman2024emcad}}}$         & \underline{89.11} & \underline{81.75} & \underline{78.65} & \underline{69.59} & 91.01 & 84.16 & \underline{86.26} & \underline{78.50} \\
MCANet$_{\textit{MIR'25 \cite{shao2025mcanet}}}$         & 89.07 & 81.69 & 76.52 & 67.10 & 90.03 & 82.49 & 85.21 & 77.09 \\
Mobileunetr$_{\textit{ECCV'24 \cite{perera2024mobileunetr}}}$   & 87.85 & 79.97 & 71.01 & 59.99 & 88.04 & 79.47 & 82.30 & 73.14 \\
MSDUNet$_{\textit{TMI'25 \cite{li2025msdunet}}}$        & 87.34 & 79.34 & 66.19 & 56.08 & 87.54 & 78.64 & 80.36 & 71.35 \\
RolllingUnet$_{\textit{AAAI'24 \cite{liu2024rolling}}}$  & 86.69 & 78.86 & 72.65 & 62.93 & 85.32 & 75.64 & 81.55 & 72.48 \\
RWKVUnet$_{\textit{arXiv'25 \cite{jiang2025rwkv}}}$     & 88.08 & 80.49 & 76.63 & 67.22 & 90.42 & 83.26 & 85.04 & 76.99 \\
Segman$_{\textit{CVPR'25 \cite{fu2025segman}}}$        & 85.67 & 77.35 & 62.95 & 52.41 & 83.47 & 72.59 & 77.36 & 67.45 \\
SwinUnet$_{\textit{ECCV'22 \cite{cao2022swin}}}$       & 87.28 & 79.48 & 68.90 & 58.20 & 85.87 & 76.26 & 80.68 & 71.31 \\
Ukan$_{\textit{AAAI'25 \cite{li2025u}}}$          & 84.39 & 75.60 & 59.20 & 48.02 & 80.07 & 67.69 & 74.55 & 63.77 \\
Unet$_{\textit{MICCAI'15 \cite{ronneberger2015u}}}$         & 85.85 & 77.54 & 68.73 & 59.25 & 86.81 & 77.43 & 80.46 & 71.41 \\
VMUnet$_{\textit{MICCAI'24 \cite{ruan2024vm}}}$      & 86.91 & 79.00 & 69.09 & 59.66 & 87.88 & 79.36 & 81.29 & 72.67 \\
\midrule
MedSAM$_{\textit{Nature Communication'24 \cite{ma2024segment}}}$ & 84.55 & 75.88 & 71.54 & 61.34 & 81.73 & 70.12 & 79.27 & 69.11 \\
SAMed$_{\textit{arXiv'23 \cite{zhang2023customized}}}$               & 87.24 & 79.39 & 73.41 & 62.87 & 87.82 & 78.95 & 82.82 & 73.74 \\
SAMmed2D$_{\textit{arXiv'23 \cite{cheng2023sammed2d}}}$          & 87.06 & 79.01 & 66.91 & 55.41 & 81.06 & 69.28 & 78.34 & 67.90 \\
SAM-Reins$_{\textit{CVPR'24\cite{wei2024stronger}}}$          & 86.94 & 78.80 & 73.66 & 64.07 & 84.83 & 74.62 & 81.81 & 72.50 \\
SAM-ConvLoRA$_{\textit{ICLR'24\cite{zhong2024convolution}}}$       & 88.37 & 80.84 & 77.26 & 68.23 & 90.86 & 83.61 & 85.50 & 77.56 \\
Medical SAM Adapter$_{\textit{MIA'25 \cite{wu2025medical}}}$ & 88.53 & 81.07 & 77.66 & 69.09 & \underline{91.35} & \underline{84.64} & 85.85 & 78.27 \\
SAM-Adapter$_{\textit{CVPR'23\cite{chen2023sam}}}$        & 86.62 & 78.45 & 72.39 & 62.51 & 84.16 & 73.51 & 81.06 & 71.49 \\
\midrule
IC-MoE                         & \textbf{89.18} & \textbf{81.80} & \textbf{79.26} & \textbf{70.40} & \textbf{91.78} & \textbf{85.23} & \textbf{86.74} & \textbf{79.14} \\
\bottomrule
\end{tabular*}
\end{table*}

\section{EXPERIMENTS AND RESULTS}
\subsection{Experimental parameters, dataset, and evaluation indices}
\noindent\textbf{Experimental setting.} First, to train the segmentation models, all the experiments were implemented within the \textit{PyTorch} framework. Second, hyperparameters were uniformly set: the batch size was adjusted to 8 or 32 on the basis of image resolution, with 100 epochs. We also fixed all random seeds to 0 to ensure reproducibility. Third, experiments were conducted on a \textit{32 GB VGPU} via the \textit{Autodl} platform. Our experimental environments are as follows: \textit{Ubuntu 20.04}, \textit{PyTorch 2.0.0}, \textit{Python 3.8}, \textit{CUDA 11.8}, CPU: \textit{Intel}® \textit{Xeon}® \textit{Platinum 8458P}. Fourth, the learning rates for all the models were set as follows: $1 \times 10^{-5}$ for epochs 1–50, $5 \times 10^{-6}$ for epochs 50–75, and $1 \times 10^{-6}$ for epochs 75–100. \textit{RMSprop} was used as the optimizer for all the models. We employed image preprocessing techniques such as scaling and normalization on the dataset via \textit{‘torchvision.transforms’}. The segmentation loss function for all the models is defined by Eq.(\ref{Eq:11}):

\begin{equation}
    {{\mathcal{L}}_{seg}}=\alpha \times {{\mathcal{L}}_{ce}}+\beta \times {{\mathcal{L}}_{dice}}
\label{Eq:11}
\end{equation}

Fifth, we evaluate the segmentation performance of the model via the following metrics: DSC, IoU, accuracy, recall, and precision.

\noindent\textbf{Datasets. (a) ISIC 2018 (International Skin Imaging Collaboration) dataset \cite{codella2019skin}.} This dataset is one of the most challenging publicly available datasets for dermatoscopy image analysis and is primarily used for skin lesion segmentation. We followed the division scheme from prior research \cite{ruan2022malunet} and obtained 1886 training images and 808 validation images. We adjusted the input resolution to 256×256 pixels.

\noindent\textbf{(b) BUSI (Breast Ultrasound Image) dataset \cite{al2020dataset}.} This is a public breast ultrasound image dataset primarily used for breast tumor segmentation. It contains three categories of images: normal, benign, and malignant. This study utilizes the benign and malignant categories, totaling 647 images.which were split into a training set of 486 images and a validation set of 161 images at a 3:1 ratio. We adjusted the input resolution to 256×256 pixels.

\noindent\textbf{(c) GLaS (Gland Segmentation in Colon Histology Images) dataset \cite{sirinukunwattana2017gland}.} This is a colon cancer histology image dataset for gland segmentation. The official division scheme contains 85 training images and 80 validation images. We adjusted the input resolution to 512×512 pixels.

\subsection{Ablation experiment}
The core components of IC-MoE are ECFM and SgCL, which enhance the transferability of high-level semantics. Moreover, they maintain the structural integrity of pretrained weights. Therefore, it is essential to conduct ablation experiments on ECFM and SgCL. We fix all the hyperparameters to ensure fairness. See \ref{tbl1}. After introducing the ECFM, the model improved by 0.2\% and 1.58\% in terms of the DSC and recall, respectively. The IoU, accuracy, and precision decreased by 0.13\%, 0.15\%, and 0.78\%, respectively. After introducing SgCL, the model improved by 1.6\%, 1.31\%, 0.12\%, and 2.84\% in terms of the DSC, IoU, accuracy, and recall, respectively, while decreasing the precision by 0.34\%. Simultaneously, we observed that the model's performance gains were relatively limited without SgCL. Therefore, we can preliminarily conclude that the introduction of SgCL facilitates intelligent communication among experts.

\subsection{Medical image segmentation experiment}
To comprehensively evaluate the generalization capabilities of IC-MoE, we conducted comparative experiments across three datasets: ISIC, BUSI, and GLAS. These datasets originate from various medical image segmentation tasks, fulfilling the requirements for multitask learning. We compared 13 traditional segmentation models with 7 popular fine-tuning models to quantitatively assess the medical image segmentation performance of IC-MoE. We find that EMCAD and Medical SAM Adapter exhibit the best overall performance among these SOTAs. Our proposed IC-MoE introduces expert modules and contrastive learning, thereby enhancing high-level features while preserving the structural integrity of pretrained weights. Table \ref{tbl2} shows that, compared with EMCAD, the IC-MoE model achieves an average improvement of 0.48\% in the DSC and 0.64\% in the IoU. Compared with the medical SAM adapter, the IC-MoE achieves average DSC and IoU improvements of 0.89\% and 0.87\%, respectively.

Taking the BUSI dataset as an example, we further evaluate the IC-MoE's superiority via violin plots; See Figure \ref{FIG:4}. Overall, the median values of the IC-MoE model are significantly higher than those of the other models. Moreover, the interval widths of the IC-MoE are concentrated in the high DSC range, with the narrowest width observed in the low DSC range. This indicates that the IC-MoE achieves the best image segmentation performance on the BUSI dataset.

\begin{figure*}[t]
    \centering
    \includegraphics[width=\linewidth,keepaspectratio]{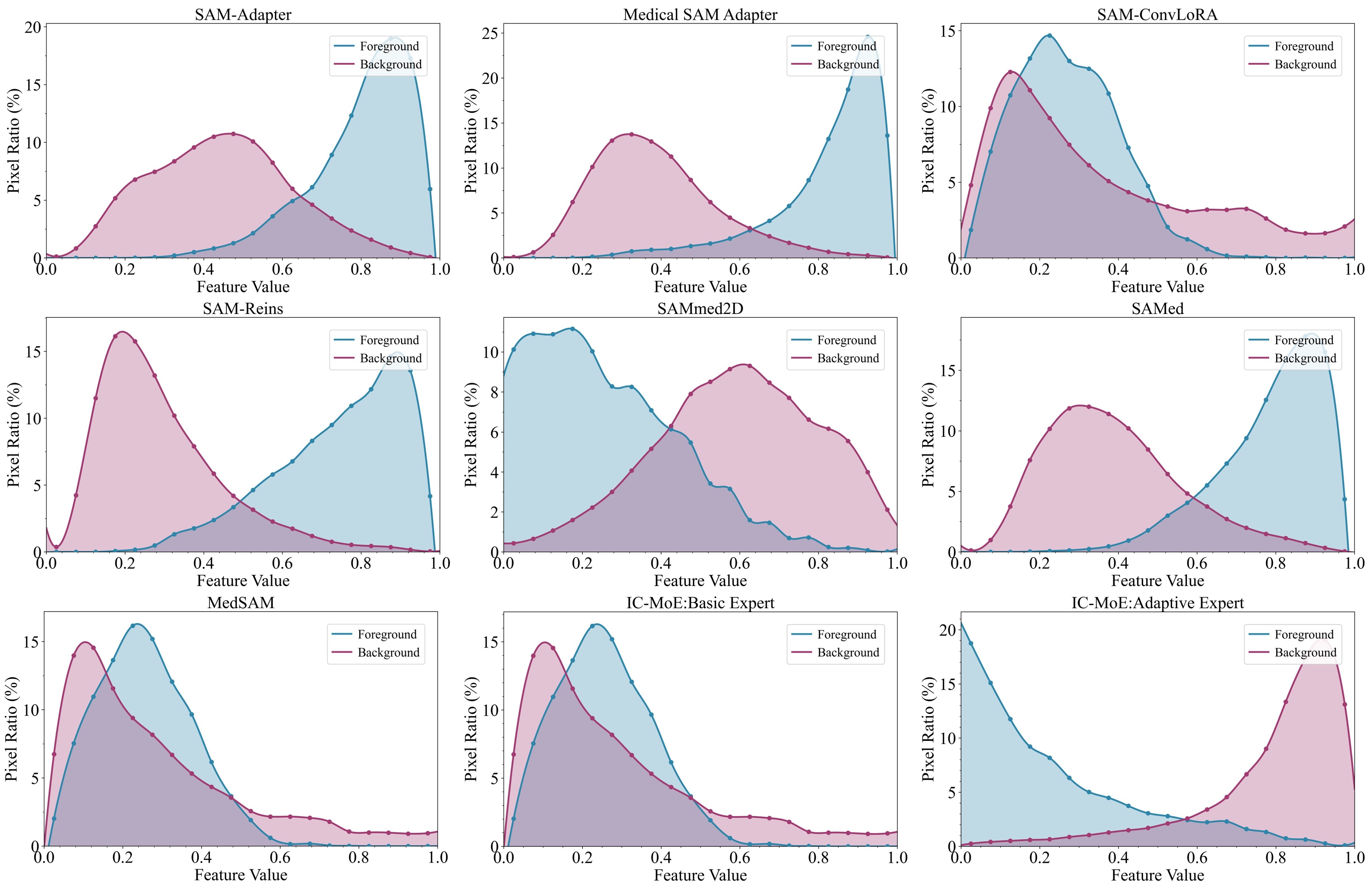}
    \caption{Feature distribution plots.}
    \label{FIG:6}
\end{figure*}

\noindent\textbf{BUSI dataset.} As shown in Table \ref{tbl3}, except for a 0.34\% decrease in precision, the IC-MoE model outperformed the fine-tuned SOTA model by 1.6\%, 1.31\%, 0.12\%, and 2.84\% in DSC, IoU, accuracy, and recall, respectively. Except for a 0.02\% decrease in accuracy, the IC-MoE model outperformed the general SOTA model by 0.61\%, 0.81\%, 0.56\%, 0.45\%, and 0.64\% in DSC, IoU, recall, and precision, respectively. These results confirm that the IC-MoE achieves the best segmentation performance on the BUSI dataset and further highlight its superior generalizability.

\noindent\textbf{ISIC dataset.} As shown in Table \ref{tbl4}, the IC-MoE model outperformed the fine-tuned SOTA model by 0.65\%, 0.73\%,
0.42\%, 0.26\%, and 0.63\% in the DSC, IoU, accuracy, recall, and precision, respectively. Except for slight decreases in accuracy and recall, the IC-MoE model outperformed the general SOTA model by 0.07\%, 0.05\%, and 0.26\% in DSC, IoU, accuracy, and precision, respectively. These results confirm that the IC-MoE achieves the best segmentation performance on the ISIC dataset and further highlight its superior generalizability.

\noindent\textbf{GLAS dataset.} As shown in Table \ref{tbl5}, except for a 0.15\% decrease in Recall, the IC-MoE model outperformed the fine-tuned SOTA model by 0.43\%, 0.59\%, 0.37\%, and 0.73\% in DSC, IoU, accuracy, and precision, respectively. The IC-MoE model outperformed the general SOTA model by 0.77\%, 1.07\%, 0.63\%, 0.03\%, and 1.32\% in terms of the DSC, IoU, accuracy, recall, and precision, respectively. These results confirm that IC-MoE achieves the best segmentation performance on the GLAS dataset and further highlight its superior generalizability.

\begin{figure}
    \centering
    \includegraphics[width=\linewidth,keepaspectratio]{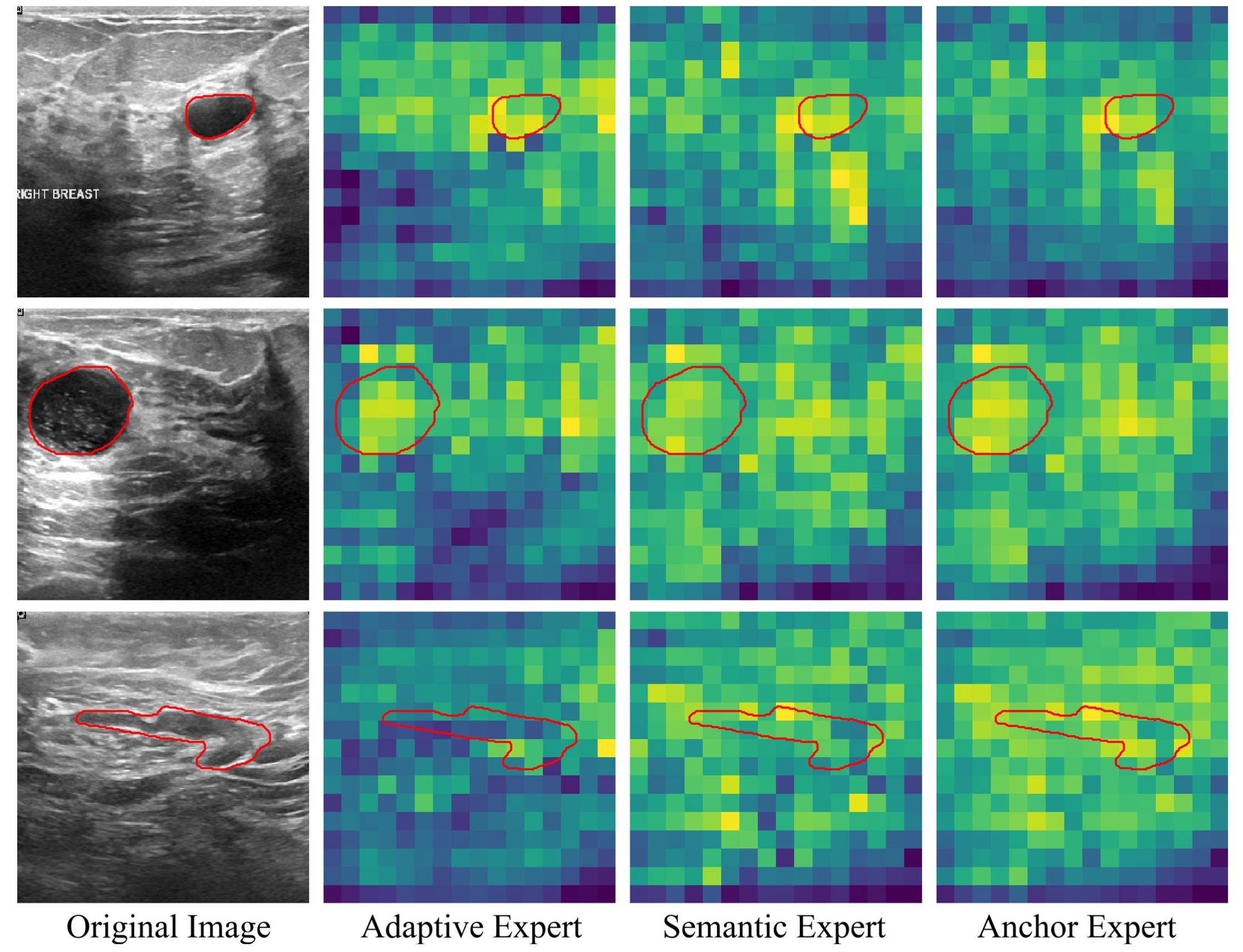}
    \caption{Visualization of features.}
    \label{FIG:5}
\end{figure}

\begin{figure*}[t]
    \centering
    \includegraphics[width=\linewidth,keepaspectratio]{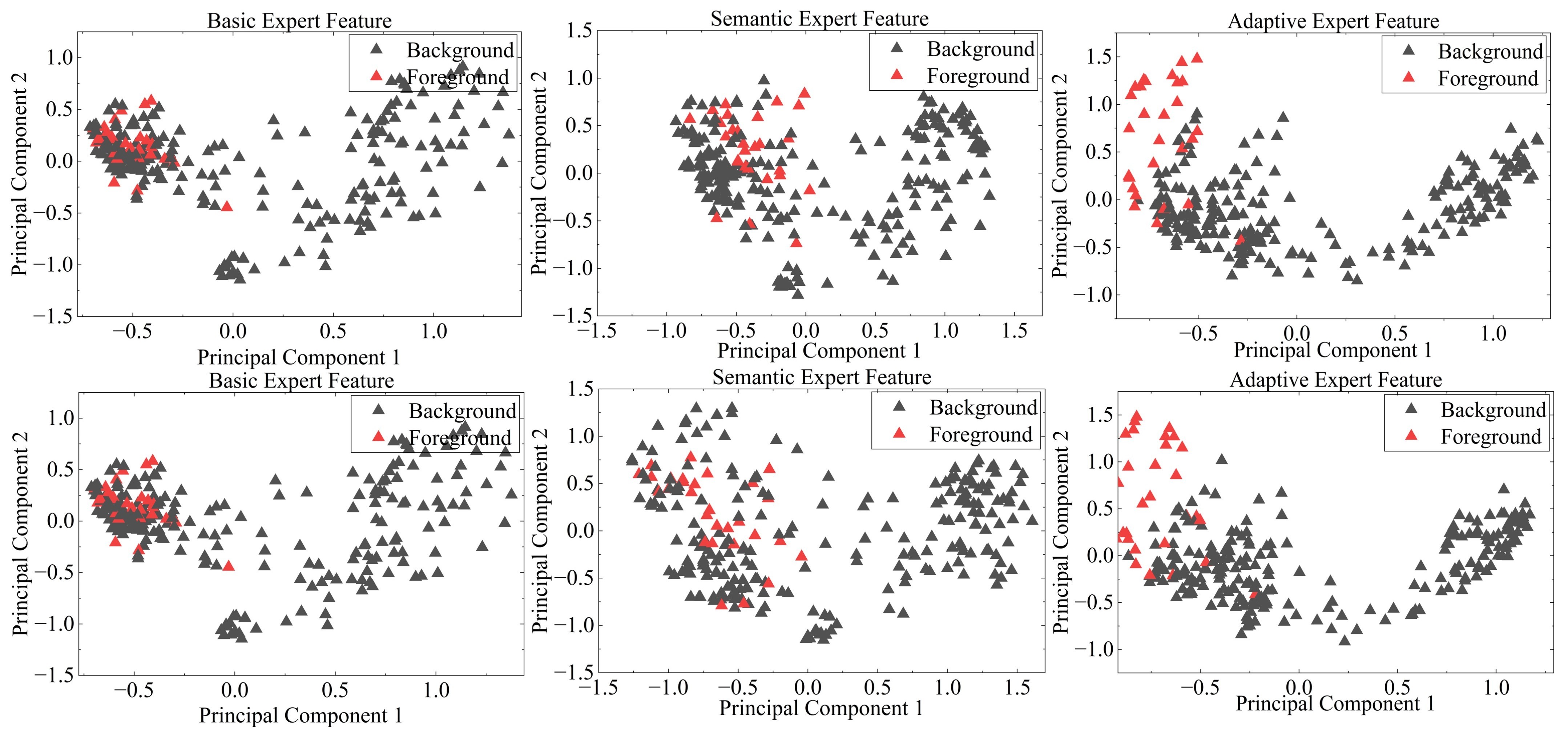}
    \caption{2D space plots of feature representations: Top 3 (with SgCL) and bottom 3 (without SgCL).}
    \label{FIG:7}
\end{figure*}

\begin{table}[width=\linewidth, cols=6, pos=t] 
\caption{Segmentation Results on the BUSI Dataset.}\label{tbl3}
\setlength{\tabcolsep}{3pt}  
\begin{tabular*}{\tblwidth}{@{} L c c c c c @{} }
\toprule
Model & \multicolumn{1}{C}{DSC} & \multicolumn{1}{C}{IoU} & \multicolumn{1}{C}{Acc} & \multicolumn{1}{C}{Recall} & \multicolumn{1}{C@{}}{Precision} \\
\midrule
AHGNN       & 72.25 & 63.02 & 95.30 & 71.26 & \textbf{80.92} \\
CGRSEG      & 66.73 & 55.57 & 94.05 & 75.23 & 66.81 \\
Emcad       & \underline{78.65} & \underline{69.59} & \textbf{95.96} & \underline{82.43} & 79.29 \\
MCANet      & 76.52 & 67.10 & 95.86 & 77.74 & \underline{80.23} \\
Mobileunetr & 71.01 & 59.99 & 95.27 & 72.51 & 76.97 \\
MSDUNet     & 66.19 & 56.08 & 94.68 & 65.22 & 74.88 \\
RolllingUnet& 72.65 & 62.93 & 95.22 & 75.51 & 76.34 \\
RWKVUnet    & 76.63 & 67.22 & 95.56 & 79.56 & 78.31 \\
Segman      & 62.95 & 52.41 & 94.48 & 62.69 & 73.86 \\
SwinUnet    & 68.90 & 58.20 & 94.41 & 68.74 & 75.37 \\
Ukan        & 59.20 & 48.02 & 93.67 & 63.03 & 65.27 \\
Unet        & 68.73 & 59.25 & 94.66 & 68.08 & 79.82 \\
VMUnet      & 69.09 & 59.66 & 94.82 & 69.75 & 75.45 \\
\midrule
MedSAM      & 71.54 & 61.34 & 94.96 & 72.64 & 76.50 \\
SAMed       & 73.41 & 62.87 & 94.43 & 81.03 & 73.14 \\
SAMmed2D    & 66.91 & 55.41 & 94.29 & 69.69 & 72.91 \\
SAM-Reins   & 73.66 & 64.07 & 95.19 & 75.03 & 77.57 \\
SAM-ConvLoRA& 77.26 & 68.23 & 95.80 & 78.67 & 80.13 \\
Medical SAM Adapter & 77.66 & 69.09 & 95.82 & 80.15 & 80.08 \\
SAM-Adapter & 72.39 & 62.51 & 95.18 & 75.45 & 75.38 \\
\midrule
IC-MoE & \textbf{79.26} & \textbf{70.40} & \underline{95.94} & \textbf{82.99} & 79.74 \\
\bottomrule
\end{tabular*}
\end{table}

\subsection{Experiments on feature representation}
Our feature representation experiments have two objectives. First, we conduct a qualitative assessment of IC-MoE's feature representation capabilities. Second, we verify whether the IC-MoE can fully represent high-level features. We preserve the feature tensor outputs of the BSAEs module and reshape them for feature visualization. To enhance readability, we marked the precise locations of the lesions in the images with red outlines. We subsequently performed principal component analysis on the feature tensors to compare the feature distributions between the foreground and background intuitively.

Figure \ref{FIG:5} presents several breast ultrasound images alongside feature heatmaps. The experiments demonstrate variations in feature extraction capabilities across experts. Taking the third sample as an example, the adaptive expert exhibits insufficient lesion recognition capability, whereas the others demonstrate strong high-level semantic feature representation. This finding indicates that the ECFM module enhances the high-level semantic representation capability of medical images while preserving the structural integrity of pretrained weights. Finally, we constructed a feature distribution map by plotting normalized PCA values on the x-axis (divided into 20 intervals) and the number of pixels
falling within each interval on the y-axis. These distribution graphs enable direct comparisons of foreground and background distributions. Compared with the other models, the IC-MoE model results in the largest difference in peak values between the foreground and background, along with overall higher peaks (see Figure \ref{FIG:6}). This suggests that IC-MoE achieves stronger foreground–background discrimination while maintaining greater intraclass feature consistency.

In summary, IC-MoE exhibits high-level semantic representation capabilities, outperforming existing SOTA models in high-level features.

\begin{figure*}
    \centering
    \includegraphics[width=\linewidth,keepaspectratio]{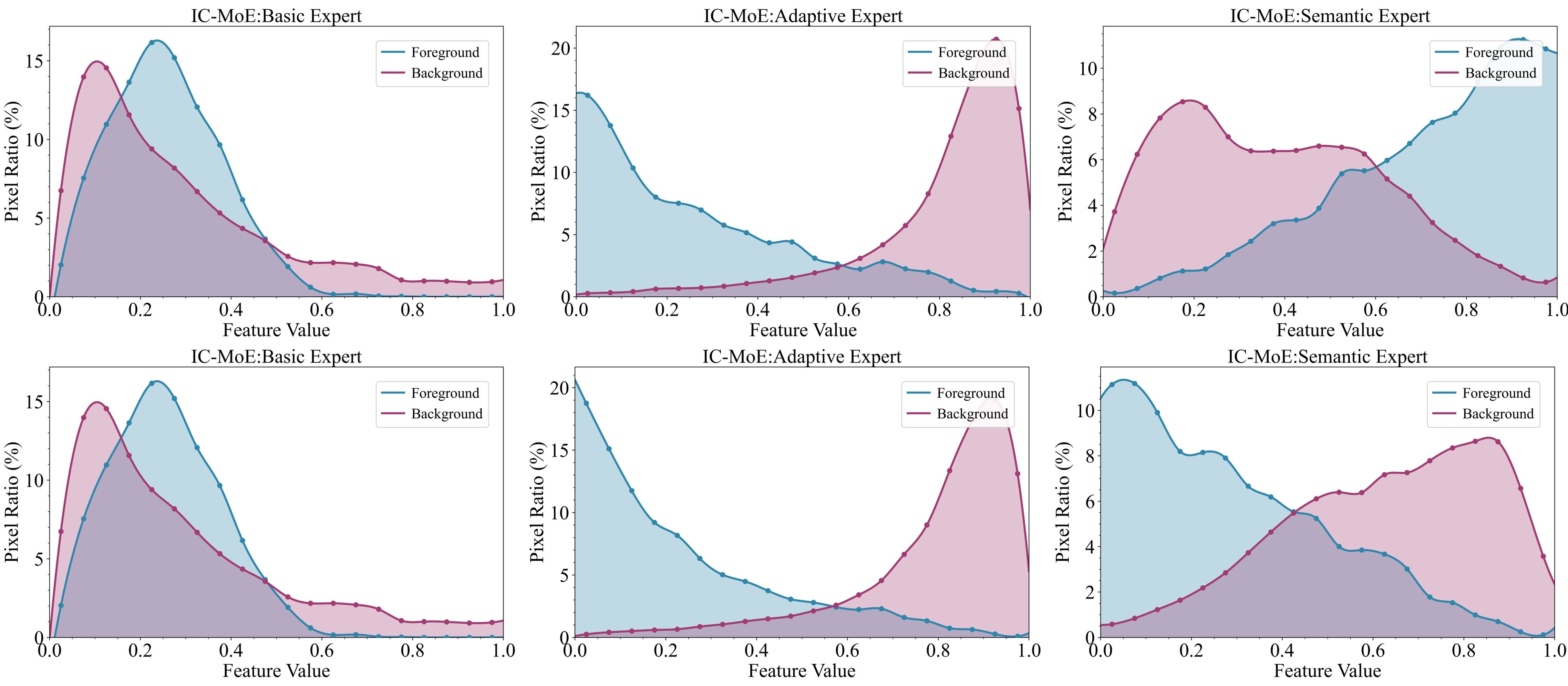}
    \caption{Feature distribution plots.}
    \label{FIG:8}
\end{figure*}

\begin{figure}
    \centering
    \includegraphics[width=\linewidth,keepaspectratio]{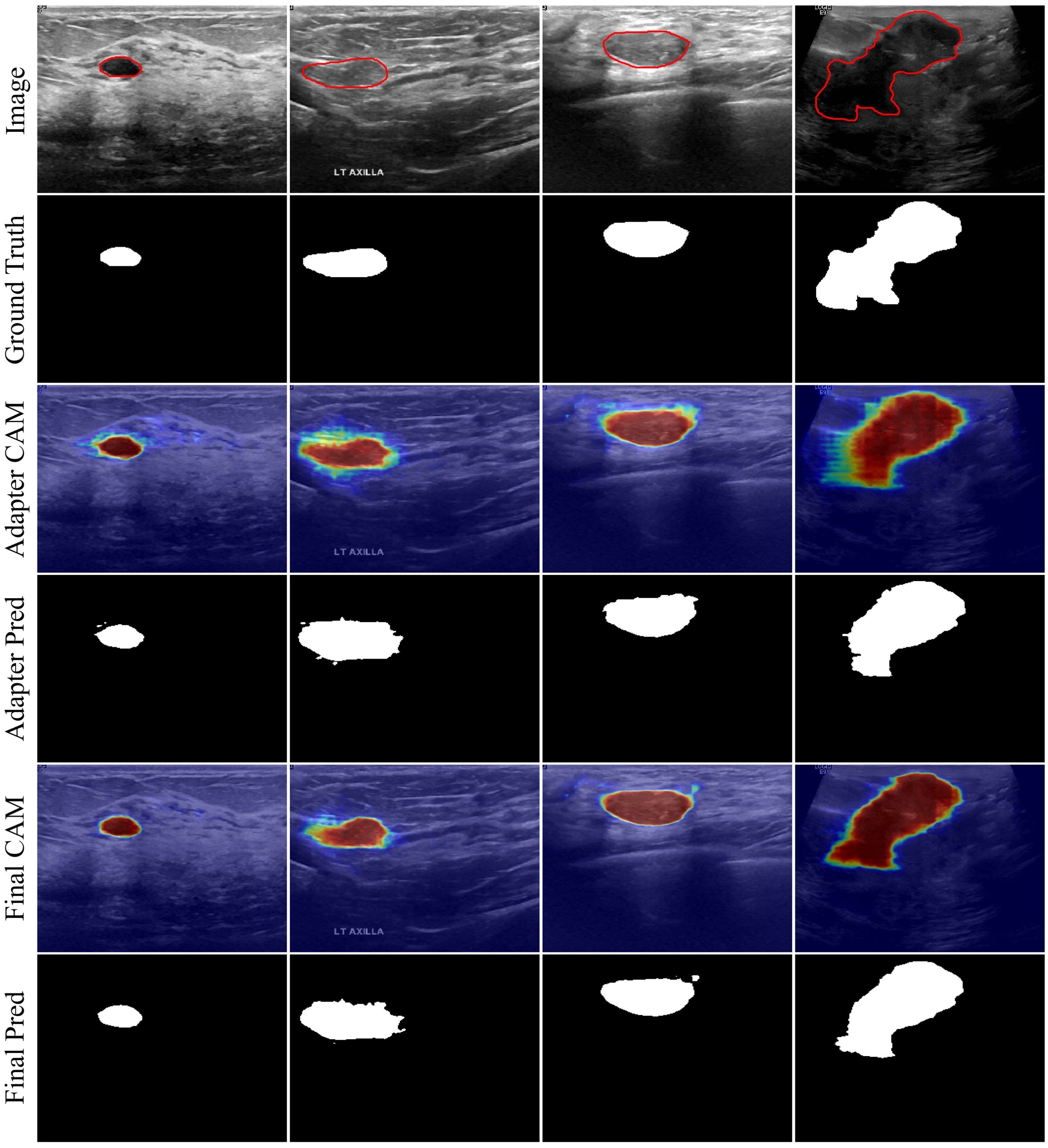}
    \caption{The performance of IC-MoE on the BUSI dataset.}
    \label{FIG:9}
\end{figure}

\subsection{Experiments on Semantic-guided Contrastive Loss}
We conducted SgCL ablation experiments with two objectives. First, we observed changes in the model's feature representation capabilities before and after introducing SgCL. Second, we validated that SgCL effectively mitigates the insufficient representation of high-level features and preserves the structural integrity of pretrained weights. Like in Section 3.4, we plotted feature distribution graphs to visually observe the effects of SgCL. As shown in Figure \ref{FIG:8}, the distance between the foreground and background features extracted by the adaptive expert further increases after the introduction of SgCL. The semantic expert exhibits relatively minor changes, but we can still observe an increase in the extracted feature distance. To explore the changes in feature distribution deeply, we perform PCA on the feature tensors. Considering the poor interpretability of analyzing all samples together, we selected one sample for comparison. The results are shown in Figure \ref{FIG:7}. Basic expert: With frozen parameters, its feature distribution remains unchanged across the two cases. Semantic expert: Its distribution range further contracts and gradually converges toward the basic expert. This indicates that the expert is learning pretrained knowledge from the basic expert. Adaptive expert: While maintaining strong feature extraction capabilities, it retains distinctiveness from other experts.

In summary, SgCL preserves the structural integrity of pretrained weights. The introduction of semantic information strengthens the supervisory signal, thereby enhancing the model’s capacity to represent high-level features.

\subsection{Experiments on visualizing the IC-MoE}
To further analyze the model's effectiveness, we visualized the final outputs of the IC-MoE model alongside those from adaptive expert fine-tuning. See Figure \ref{FIG:9}. First, both fine-tuning approaches accurately segmented the tumor regions. However, the IC-MoE method achieved more precise tumor segmentation than the prefusion outputs did. Additionally, IC-MoE demonstrated stronger resistance to artifacts and more adequate extraction of high-level features of tumors.

Furthermore, we employed Grad-CAM to generate activation maps for visualization. Grad-CAM heatmaps reflect the model's attention distribution: red areas indicate higher probabilities of tumor presence, whereas blue areas indicate lower probabilities. The heatmaps further highlight the outstanding performance of the IC-MoE model. IC-MoE demonstrates higher confidence in lesion detection and more Precise segmentation of lesion edges. This finding indicates that IC-MoE captures more comprehensive high-level features related to tumors.

\begin{figure*}
    \centering
    \includegraphics[width=\linewidth,keepaspectratio]{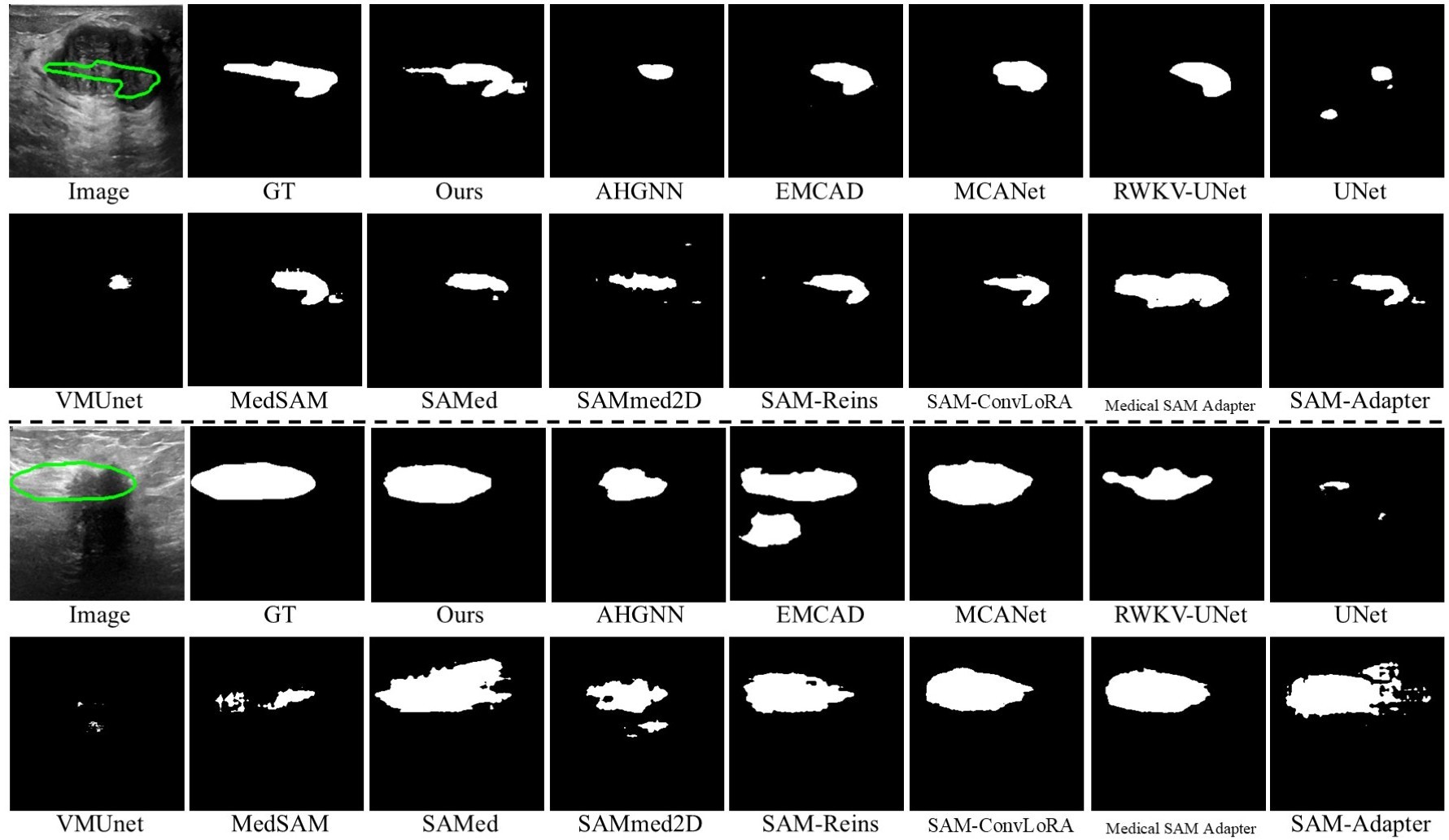}
    \caption{Model performance on the BUSI dataset.}
    \label{FIG:10}
\end{figure*}

\begin{table}[width=\linewidth, cols=6, pos=t]
\caption{Segmentation Results on the ISIC Dataset.}\label{tbl4}
\setlength{\tabcolsep}{3pt}  
\begin{tabular*}{\tblwidth}{@{} L c c c c c @{} }
\toprule
Model & \multicolumn{1}{C}{DSC} & \multicolumn{1}{C}{IoU} & \multicolumn{1}{C}{Acc} & \multicolumn{1}{C}{Recall} & \multicolumn{1}{C@{}}{Precision} \\
\midrule
AHGNN       & 86.99 & 79.17 & 93.94 & 86.99 & \textbf{91.23} \\
CGRSEG      & 86.95 & 78.67 & 94.39 & 89.30 & 88.39 \\
Emcad       & \underline{89.11} & \underline{81.75} & \textbf{95.09} & \underline{91.52} & 89.58 \\
MCANet      & 89.07 & 81.69 & \textbf{95.09} & \textbf{92.21} & 88.96 \\
Mobileunetr & 87.85 & 79.97 & 94.68 & 91.22 & 88.00 \\
MSDUNet     & 87.34 & 79.34 & 94.66 & 88.57 & 89.96 \\
RolllingUnet& 86.69 & 78.86 & 93.91 & 87.88 & 89.92 \\
RWKVUnet    & 88.08 & 80.49 & 94.75 & 91.23 & 88.44 \\
Segman      & 85.67 & 77.35 & 93.77 & 86.73 & 89.51 \\
SwinUnet    & 87.28 & 79.48 & 94.30 & 89.30 & 89.09 \\
Ukan        & 84.39 & 75.60 & 93.20 & 86.58 & 87.59 \\
Unet        & 85.85 & 77.54 & 93.30 & 86.55 & 89.44 \\
VMUnet      & 86.91 & 79.00 & 94.11 & 90.34 & 87.73 \\
\midrule
MedSAM      & 84.55 & 75.88 & 92.97 & 87.38 & 87.45 \\
SAMed       & 87.24 & 79.39 & 93.99 & 90.01 & 88.80 \\
SAMmed2D    & 87.06 & 79.01 & 94.23 & 87.94 & \underline{90.35} \\
SAM-Reins   & 86.94 & 78.80 & 93.95 & 90.12 & 87.92 \\
SAM-ConvLoRA& 88.37 & 80.84 & 94.52 & 90.57 & 89.75 \\
Medical SAM Adapter & 88.53 & 81.07 & 94.56 & 91.24 & 89.21 \\
SAM-Adapter & 86.62 & 78.45 & 93.82 & 89.3 & 88.22 \\
\midrule
IC-MoE & \textbf{89.18} & \textbf{81.80} & \underline{94.98} & 91.50 & 89.84 \\
\bottomrule
\end{tabular*}
\end{table}

\begin{figure*}
    \centering
    \includegraphics[width=\linewidth,keepaspectratio]{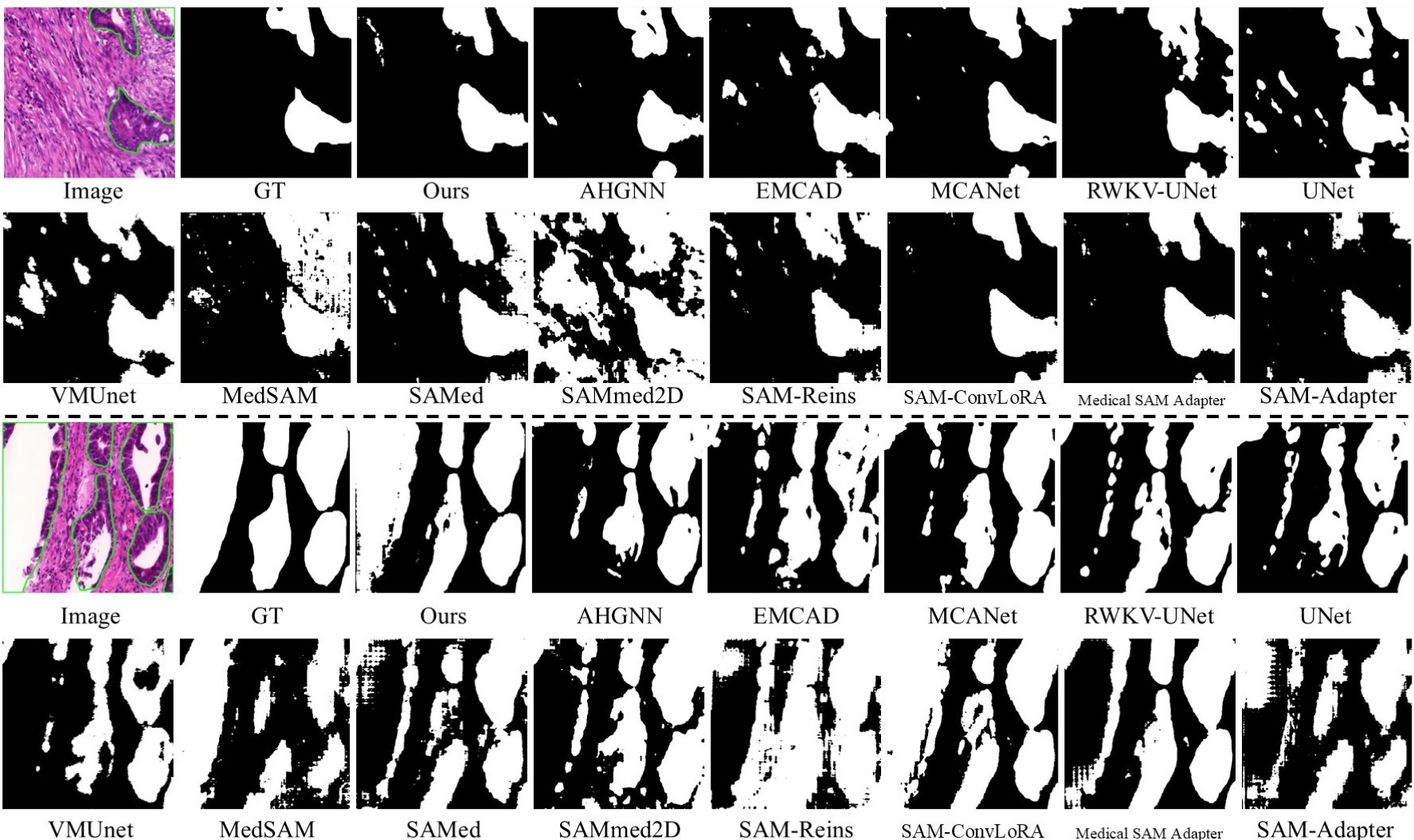}
    \caption{Model performance on the GLAS dataset.}
    \label{FIG:11}
\end{figure*}

\begin{table}[width=\linewidth, cols=6, pos=t]
\caption{Segmentation Results on the GLAS Dataset.}\label{tbl5}
\setlength{\tabcolsep}{3pt}  
\begin{tabular*}{\tblwidth}{@{} L c c c c c @{} }
\toprule
Model & \multicolumn{1}{C}{DSC} & \multicolumn{1}{C}{IoU} & \multicolumn{1}{C}{Acc} & \multicolumn{1}{C}{Recall} & \multicolumn{1}{C@{}}{Precision} \\
\midrule
AHGNN       & 89.17 & 81.16 & 89.41 & 88.72 & \underline{90.81} \\
CGRSEG      & 84.57 & 74.97 & 83.50 & 91.22 & 82.11 \\
Emcad       & 91.01 & 84.16 & 91.50 & 92.86 & 90.00 \\
MCANet      & 90.03 & 82.49 & 90.32 & 91.11 & 89.81 \\
Mobileunetr & 88.04 & 79.47 & 88.62 & 88.30 & 88.99 \\
MSDUNet     & 87.54 & 78.64 & 88.15 & 89.68 & 86.75 \\
RolllingUnet& 85.32 & 75.64 & 86.01 & 87.62 & 85.50 \\
RWKVUnet    & 90.42 & 83.26 & 90.89 & 92.46 & 89.66 \\
Segman      & 83.47 & 72.59 & 83.08 & 87.73 & 81.12 \\
SwinUnet    & 85.87 & 76.26 & 86.78 & 88.03 & 85.54 \\
Ukan        & 80.07 & 67.69 & 80.35 & 84.48 & 78.28 \\
Unet        & 86.81 & 77.43 & 87.27 & 88.01 & 87.17 \\
VMUnet      & 87.88 & 79.36 & 88.66 & 89.00 & 88.49 \\
\midrule
MedSAM      & 81.73 & 70.12 & 81.86 & 84.33 & 81.38 \\
SAMed       & 87.82 & 78.95 & 88.15 & 88.82 & 88.37 \\
SAMmed2D    & 81.06 & 69.28 & 82.02 & 83.94 & 81.92 \\
SAM-Reins   & 84.83 & 74.62 & 84.73 & 87.89 & 83.34 \\
SAM-ConvLoRA& 90.86 & 83.61 & 91.04 & 91.51 & 90.89 \\
Medical SAM Adapter & \underline{91.35} & \underline{84.64} & \underline{91.76} & \textbf{93.04} & 90.59 \\
SAM-Adapter & 84.16 & 73.51 & 84.81 & 85.06 & 85.28 \\
\midrule
IC-MoE & \textbf{91.78} & \textbf{85.23} & \textbf{92.13} & \underline{92.89} & \textbf{91.32} \\
\bottomrule
\end{tabular*}
\end{table}

\section{DISCUSSION AND ANALYSIS}
\subsection{Discussion of the IC-MoE}
According to the introduction, conventional fine-tuning methods generally have two main problems. They lack the ability to represent high-level features for medical images, thereby compromising the structural integrity of pretrained weights. To address these problems, we propose the IC-MoE model. We introduce an expert module and a pixel probability adaptive voting strategy. Furthermore, we introduce a semantic-guided contrastive learning approach. This enables the model to supplement the high-level features while preserving the structural integrity of pretrained weights. As demonstrated in Tables \ref{tbl1}–\ref{tbl5} and Figures \ref{FIG:3}–\ref{FIG:9}, the IC-MoE outperforms 20 other SOTA models in both quantitative metrics and qualitative analysis. Therefore, we can assert that IC-MoE successfully integrates high-level features. Moreover, it maintains the structural integrity of pretrained weights, thereby enhancing the capabilities of SAM models in medical image segmentation.

\begin{figure*}
    \centering
    \includegraphics[width=\linewidth,keepaspectratio]{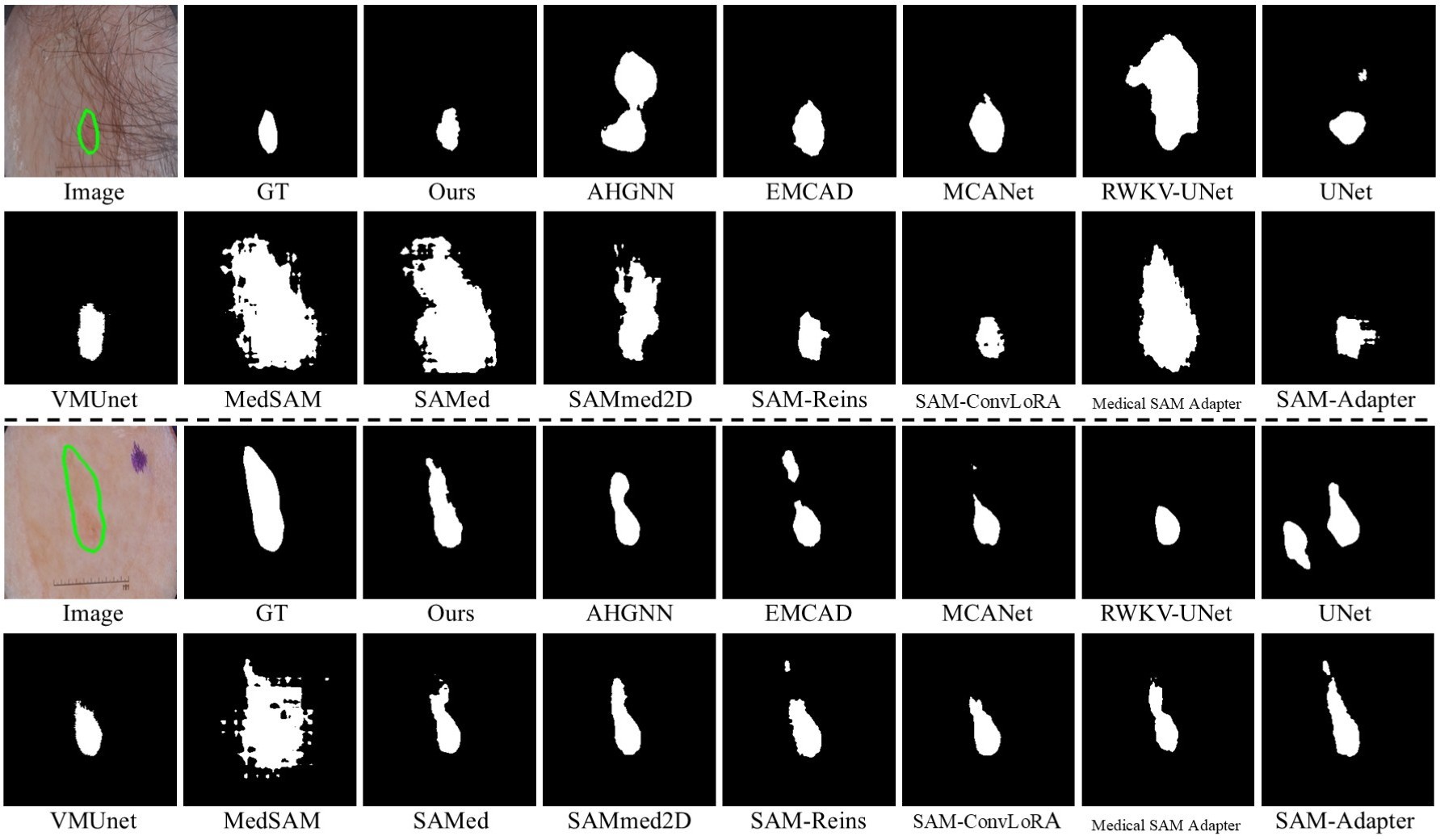}
    \caption{Model performance on the ISIC dataset.}
    \label{FIG:12}
\end{figure*}

\begin{table}[width=\linewidth, cols=5, pos=t]
\caption{Model Complexity Comparison.}
\label{tbl6}
\setlength{\tabcolsep}{3pt}  
\begin{tabular*}{\tblwidth}{@{} L c c c c @{} }
\toprule
Model &
\makecell[c]{Params\\(M)} &
\makecell[c]{Trainable\\Params (M)} &
\makecell[c]{GFLOPs\\(G)} &
\makecell[c]{Time\\(ms)} \\
\midrule
AHGNN       & 43.08 & 43.08 & 37.34 & 7.59 \\
CGRSEG      & 19.08 & 19.08 & 1.88 & 11.38 \\
Emcad       & 26.77 & 26.77 & 5.60 & 8.77 \\
MCANet      & 26.83 & 26.83 & 9.11 & 9.58 \\
Mobileunetr & 14.16 & 14.16 & 5.80 & 11.40 \\
MSDUNet     & 22.17 & 22.17 & 6.45 & 37.70 \\
RolllingUnet& 28.32 & 28.32 & 32.87 & 57.59 \\
RWKVUnet    & 17.37 & 17.37 & 14.58 & 7.30 \\
Segman      & 29.40 & 29.40 & 5.69 & 22.60 \\
SwinUnet    & 41.39 & 41.39 & 11.37 & 8.30 \\
Ukan        & 6.36 & 6.36 & 1.75 & 7.16 \\
Unet        & 17.26 & 17.26 & 40.19 & 3.19 \\
VMUnet      & 27.43 & 27.43 & 4.11 & 7.27 \\
\midrule
MedSAM      & 90.46 & 3.78 & 32.27 & 10.45 \\
SAMed       & 90.60 & 3.93 & 32.36 & 10.34 \\
SAMmed2D    & 270.96 & 184.29 & 65.36 & 14.43 \\
SAM-Reins   & 92.09 & 5.41 & 33.11 & 10.61 \\
SAM-ConvLoRA& 90.58 & 3.90 & 19.38 & 20.96 \\
Medical SAM Adapter & 97.56 & 10.88 & 35.33 & 11.04 \\
SAM-Adapter & 90.52 & 3.85 & 32.34 & 10.98 \\
\midrule
IC-MoE & 279.06 & 33.23 & 99.88 & 30.78 \\
\bottomrule
\end{tabular*}
\end{table}

\subsection{Discussion of visual segmentation results}
By experiments and results, as described in section 3, we quantitatively evaluated the segmentation performance of the models across multiple datasets. To provide a more intuitive comparison of segmentation capabilities among different models, this section further presents visual comparisons of some prediction results. As shown in Figures \ref{FIG:10}–\ref{FIG:12}, our proposed IC-MoE model achieves greater precision in locating lesion regions. This qualitative analysis not only aligns with the quantitative experimental conclusions but also further demonstrates that IC-MoE possesses more high-level features and stronger generalization performance in medical image segmentation tasks.

\subsection{Discussion of model complexity}
To analyze model complexity, we conducted a comparative analysis of parameters, trainable parameters, computational complexity, and average inference time for each model on the BUSI dataset. As shown in Table \ref{tbl6}, our IC-MoE model has the highest number of parameters and computational complexity. Compared with SAMmed2D, the IC-MoE model has fewer trainable parameters while achieving a significantly higher DSC. Our IC-MoE model's DSC exceeds those of all other models, indicating that it has the strongest feature representation capability. Furthermore, despite higher overall training costs, the IC-MoE model achieves an average inference time of only 30.78 ms per image. This inference speed falls well within the acceptable range for clinical diagnostic applications. In summary, the IC-MoE demonstrates superior feature representation capabilities. Despite the higher training cost, this fine-tuning approach holds reasonable potential for enhancing clinical diagnostic capabilities.

\section{CONCLUSIONS AND FUTURE WORK} 
\noindent\textbf{Conclusion.} Parameter-efficient fine-tuning is a general approach for knowledge transfer in large models; however, existing methods suffer from two main limitations: insufficient high-level features and compromised structural integrity of pretrained weights. To address these two problems, we propose a novel segmentation model called IC-MoE. First, we construct three core experts: the basic expert, the semantic expert, and the adaptive expert. We fuse and output these experts through a pixel probability adaptive voting strategy. This framework effectively preserves the structural integrity of pretrained weights while augmenting high-level features. Furthermore, we introduce a semantic-guided contrastive learning approach to address the issue of insufficient supervision in contrastive learning. Through these methods, the IC-MoE model significantly enhances high-level features while retaining the structural integrity of pretrained weights. Extensive experiments demonstrate that the IC-MoE outperforms existing traditional models and parameter-efficient fine-tuning methods in segmentation tasks.

\noindent\textbf{Future work.} First, we aim to further reduce the complexity and computational cost of existing expert models. Second, we adopt more advanced metric learning methods to measure distances accurately between experts in feature space. Third, we integrate the latest contrastive learning strategies to enhance complementary communication among experts.

\FloatBarrier



\printcredits

\bibliographystyle{model1-num-names}

\bibliography{cas-refs}

\begin{thebibliography}{40}
\expandafter\ifx\csname natexlab\endcsname\relax\def\natexlab#1{#1}\fi
\providecommand{\url}[1]{\texttt{#1}}
\providecommand{\href}[2]{#2}
\providecommand{\path}[1]{#1}
\providecommand{\DOIprefix}{doi:}
\providecommand{\ArXivprefix}{arXiv:}
\providecommand{\URLprefix}{URL: }
\providecommand{\Pubmedprefix}{pmid:}
\providecommand{\doi}[1]{\href{http://dx.doi.org/#1}{\path{#1}}}
\providecommand{\Pubmed}[1]{\href{pmid:#1}{\path{#1}}}
\providecommand{\bibinfo}[2]{#2}
\ifx\xfnm\relax \def\xfnm[#1]{\unskip,\space#1}\fi
\bibitem[{De~Fauw et~al.(2018)De~Fauw, Ledsam, Romera-Paredes, Nikolov, Tomasev, Blackwell, Askham, Glorot, O’Donoghue, Visentin et~al.}]{de2018clinically}
\bibinfo{author}{J.~De~Fauw}, \bibinfo{author}{J.~R. Ledsam}, \bibinfo{author}{B.~Romera-Paredes}, \bibinfo{author}{S.~Nikolov}, \bibinfo{author}{N.~Tomasev}, \bibinfo{author}{S.~Blackwell}, \bibinfo{author}{H.~Askham}, \bibinfo{author}{X.~Glorot}, \bibinfo{author}{B.~O’Donoghue}, \bibinfo{author}{D.~Visentin}, et~al.,
\newblock \bibinfo{title}{Clinically applicable deep learning for diagnosis and referral in retinal disease},
\newblock \bibinfo{journal}{Nature medicine} \bibinfo{volume}{24} (\bibinfo{year}{2018}) \bibinfo{pages}{1342--1350}.
\bibitem[{Ouyang et~al.(2020)Ouyang, He, Ghorbani, Yuan, Ebinger, Langlotz, Heidenreich, Harrington, Liang, Ashley et~al.}]{ouyang2020video}
\bibinfo{author}{D.~Ouyang}, \bibinfo{author}{B.~He}, \bibinfo{author}{A.~Ghorbani}, \bibinfo{author}{N.~Yuan}, \bibinfo{author}{J.~Ebinger}, \bibinfo{author}{C.~P. Langlotz}, \bibinfo{author}{P.~A. Heidenreich}, \bibinfo{author}{R.~A. Harrington}, \bibinfo{author}{D.~H. Liang}, \bibinfo{author}{E.~A. Ashley}, et~al.,
\newblock \bibinfo{title}{Video-based ai for beat-to-beat assessment of cardiac function},
\newblock \bibinfo{journal}{Nature} \bibinfo{volume}{580} (\bibinfo{year}{2020}) \bibinfo{pages}{252--256}.
\bibitem[{Ronneberger et~al.(2015)Ronneberger, Fischer, and Brox}]{ronneberger2015u}
\bibinfo{author}{O.~Ronneberger}, \bibinfo{author}{P.~Fischer}, \bibinfo{author}{T.~Brox},
\newblock \bibinfo{title}{U-net: Convolutional networks for biomedical image segmentation},
\newblock in: \bibinfo{booktitle}{International Conference on Medical image computing and computer-assisted intervention}, \bibinfo{organization}{Springer}, \bibinfo{year}{2015}, pp. \bibinfo{pages}{234--241}.
\bibitem[{Cao et~al.(2022)Cao, Wang, Chen, Jiang, Zhang, Tian, and Wang}]{cao2022swin}
\bibinfo{author}{H.~Cao}, \bibinfo{author}{Y.~Wang}, \bibinfo{author}{J.~Chen}, \bibinfo{author}{D.~Jiang}, \bibinfo{author}{X.~Zhang}, \bibinfo{author}{Q.~Tian}, \bibinfo{author}{M.~Wang},
\newblock \bibinfo{title}{Swin-unet: Unet-like pure transformer for medical image segmentation},
\newblock in: \bibinfo{booktitle}{European conference on computer vision}, \bibinfo{organization}{Springer}, \bibinfo{year}{2022}, pp. \bibinfo{pages}{205--218}.
\bibitem[{Isensee et~al.(2021)Isensee, Jaeger, Kohl, Petersen, and Maier-Hein}]{isensee2021nnu}
\bibinfo{author}{F.~Isensee}, \bibinfo{author}{P.~F. Jaeger}, \bibinfo{author}{S.~A. Kohl}, \bibinfo{author}{J.~Petersen}, \bibinfo{author}{K.~H. Maier-Hein},
\newblock \bibinfo{title}{nnu-net: a self-configuring method for deep learning-based biomedical image segmentation},
\newblock \bibinfo{journal}{Nature methods} \bibinfo{volume}{18} (\bibinfo{year}{2021}) \bibinfo{pages}{203--211}.
\bibitem[{Chen et~al.(2021)Chen, Lu, Yu, Luo, Adeli, Wang, Lu, Yuille, and Zhou}]{chen2021transunettransformersmakestrong}
\bibinfo{author}{J.~Chen}, \bibinfo{author}{Y.~Lu}, \bibinfo{author}{Q.~Yu}, \bibinfo{author}{X.~Luo}, \bibinfo{author}{E.~Adeli}, \bibinfo{author}{Y.~Wang}, \bibinfo{author}{L.~Lu}, \bibinfo{author}{A.~L. Yuille}, \bibinfo{author}{Y.~Zhou}, \bibinfo{title}{Transunet: Transformers make strong encoders for medical image segmentation}, \bibinfo{year}{2021}. \URLprefix \url{https://arxiv.org/abs/2102.04306}. \href{http://arxiv.org/abs/2102.04306}{\tt arXiv:2102.04306}.
\bibitem[{Caron et~al.(2021)Caron, Touvron, Misra, J{\'e}gou, Mairal, Bojanowski, and Joulin}]{caron2021emerging}
\bibinfo{author}{M.~Caron}, \bibinfo{author}{H.~Touvron}, \bibinfo{author}{I.~Misra}, \bibinfo{author}{H.~J{\'e}gou}, \bibinfo{author}{J.~Mairal}, \bibinfo{author}{P.~Bojanowski}, \bibinfo{author}{A.~Joulin},
\newblock \bibinfo{title}{Emerging properties in self-supervised vision transformers},
\newblock in: \bibinfo{booktitle}{Proceedings of the IEEE/CVF international conference on computer vision}, \bibinfo{year}{2021}, pp. \bibinfo{pages}{9650--9660}.
\bibitem[{Radford et~al.(2021)Radford, Kim, Hallacy, Ramesh, Goh, Agarwal, Sastry, Askell, Mishkin, Clark et~al.}]{radford2021learning}
\bibinfo{author}{A.~Radford}, \bibinfo{author}{J.~W. Kim}, \bibinfo{author}{C.~Hallacy}, \bibinfo{author}{A.~Ramesh}, \bibinfo{author}{G.~Goh}, \bibinfo{author}{S.~Agarwal}, \bibinfo{author}{G.~Sastry}, \bibinfo{author}{A.~Askell}, \bibinfo{author}{P.~Mishkin}, \bibinfo{author}{J.~Clark}, et~al.,
\newblock \bibinfo{title}{Learning transferable visual models from natural language supervision},
\newblock in: \bibinfo{booktitle}{International conference on machine learning}, \bibinfo{organization}{PmLR}, \bibinfo{year}{2021}, pp. \bibinfo{pages}{8748--8763}.
\bibitem[{Kirillov et~al.(2023)Kirillov, Mintun, Ravi, Mao, Rolland, Gustafson, Xiao, Whitehead, Berg, Lo et~al.}]{kirillov2023segment}
\bibinfo{author}{A.~Kirillov}, \bibinfo{author}{E.~Mintun}, \bibinfo{author}{N.~Ravi}, \bibinfo{author}{H.~Mao}, \bibinfo{author}{C.~Rolland}, \bibinfo{author}{L.~Gustafson}, \bibinfo{author}{T.~Xiao}, \bibinfo{author}{S.~Whitehead}, \bibinfo{author}{A.~C. Berg}, \bibinfo{author}{W.-Y. Lo}, et~al.,
\newblock \bibinfo{title}{Segment anything},
\newblock in: \bibinfo{booktitle}{Proceedings of the IEEE/CVF international conference on computer vision}, \bibinfo{year}{2023}, pp. \bibinfo{pages}{4015--4026}.
\bibitem[{Wang et~al.(2023)Wang, Zhang, Cao, Wang, Shen, and Huang}]{wang2023seggpt}
\bibinfo{author}{X.~Wang}, \bibinfo{author}{X.~Zhang}, \bibinfo{author}{Y.~Cao}, \bibinfo{author}{W.~Wang}, \bibinfo{author}{C.~Shen}, \bibinfo{author}{T.~Huang},
\newblock \bibinfo{title}{Seggpt: Segmenting everything in context},
\newblock \bibinfo{journal}{arXiv preprint arXiv:2304.03284}  (\bibinfo{year}{2023}).
\bibitem[{He et~al.(2023)He, Bao, Li, Grant, and Ou}]{he2023accuracy}
\bibinfo{author}{S.~He}, \bibinfo{author}{R.~Bao}, \bibinfo{author}{J.~Li}, \bibinfo{author}{P.~E. Grant}, \bibinfo{author}{Y.~Ou},
\newblock \bibinfo{title}{Accuracy of segment-anything model (sam) in medical image segmentation tasks},
\newblock \bibinfo{journal}{CoRR}  (\bibinfo{year}{2023}).
\bibitem[{Gu et~al.(2024)Gu, Dong, Yang, and Mazurowski}]{gu2024build}
\bibinfo{author}{H.~Gu}, \bibinfo{author}{H.~Dong}, \bibinfo{author}{J.~Yang}, \bibinfo{author}{M.~A. Mazurowski},
\newblock \bibinfo{title}{How to build the best medical image segmentation algorithm using foundation models: a comprehensive empirical study with segment anything model},
\newblock \bibinfo{journal}{arXiv preprint arXiv:2404.09957}  (\bibinfo{year}{2024}).
\bibitem[{Zhang and Liu(2023)}]{zhang2023customized}
\bibinfo{author}{K.~Zhang}, \bibinfo{author}{D.~Liu},
\newblock \bibinfo{title}{Customized segment anything model for medical image segmentation},
\newblock \bibinfo{journal}{arXiv preprint arXiv:2304.13785}  (\bibinfo{year}{2023}).
\bibitem[{Wu et~al.(2025)Wu, Wang, Hong, Ji, Fu, Xu, Xu, and Jin}]{wu2025medical}
\bibinfo{author}{J.~Wu}, \bibinfo{author}{Z.~Wang}, \bibinfo{author}{M.~Hong}, \bibinfo{author}{W.~Ji}, \bibinfo{author}{H.~Fu}, \bibinfo{author}{Y.~Xu}, \bibinfo{author}{M.~Xu}, \bibinfo{author}{Y.~Jin},
\newblock \bibinfo{title}{Medical sam adapter: Adapting segment anything model for medical image segmentation},
\newblock \bibinfo{journal}{Medical image analysis} \bibinfo{volume}{102} (\bibinfo{year}{2025}) \bibinfo{pages}{103547}.
\bibitem[{Cheng et~al.(2023)Cheng, Ye, Deng, Chen, Li, Wang, Su, Huang, Chen, Jiang, Sun, He, Zhang, Zhu, and Qiao}]{cheng2023sammed2d}
\bibinfo{author}{J.~Cheng}, \bibinfo{author}{J.~Ye}, \bibinfo{author}{Z.~Deng}, \bibinfo{author}{J.~Chen}, \bibinfo{author}{T.~Li}, \bibinfo{author}{H.~Wang}, \bibinfo{author}{Y.~Su}, \bibinfo{author}{Z.~Huang}, \bibinfo{author}{J.~Chen}, \bibinfo{author}{L.~Jiang}, \bibinfo{author}{H.~Sun}, \bibinfo{author}{J.~He}, \bibinfo{author}{S.~Zhang}, \bibinfo{author}{M.~Zhu}, \bibinfo{author}{Y.~Qiao}, \bibinfo{title}{Sam-med2d}, \bibinfo{year}{2023}. \URLprefix \url{https://arxiv.org/abs/2308.16184}. \href{http://arxiv.org/abs/2308.16184}{\tt arXiv:2308.16184}.
\bibitem[{Feng et~al.(2025)Feng, Zhang, Chen, Shi, Liu, Sun, Du, and Chen}]{feng2025swinsam}
\bibinfo{author}{Z.~Feng}, \bibinfo{author}{Y.~Zhang}, \bibinfo{author}{Y.~Chen}, \bibinfo{author}{Y.~Shi}, \bibinfo{author}{Y.~Liu}, \bibinfo{author}{W.~Sun}, \bibinfo{author}{L.~Du}, \bibinfo{author}{D.~Chen},
\newblock \bibinfo{title}{Swinsam: Fine-grained polyp segmentation in colonoscopy images via segment anything model integrated with a swin transformer decoder},
\newblock \bibinfo{journal}{Biomedical Signal Processing and Control} \bibinfo{volume}{100} (\bibinfo{year}{2025}) \bibinfo{pages}{107055}.
\bibitem[{Chen et~al.(2023)Chen, Zhu, Ding, Cao, Zhang, Wang, Li, Sun, Mao, and Zang}]{chen2023sam}
\bibinfo{author}{T.~Chen}, \bibinfo{author}{L.~Zhu}, \bibinfo{author}{C.~Ding}, \bibinfo{author}{R.~Cao}, \bibinfo{author}{S.~Zhang}, \bibinfo{author}{Y.~Wang}, \bibinfo{author}{Z.~Li}, \bibinfo{author}{L.~Sun}, \bibinfo{author}{P.~Mao}, \bibinfo{author}{Y.~Zang},
\newblock \bibinfo{title}{Sam fails to segment anything?--sam-adapter: Adapting sam in underperformed scenes: Camouflage, shadow, and more},
\newblock \bibinfo{journal}{arXiv preprint arXiv:2304.09148} \bibinfo{volume}{2} (\bibinfo{year}{2023}) \bibinfo{pages}{7}.
\bibitem[{Zhong et~al.(2024)Zhong, Tang, He, Fang, and Yuan}]{zhong2024convolution}
\bibinfo{author}{Z.~Zhong}, \bibinfo{author}{Z.~Tang}, \bibinfo{author}{T.~He}, \bibinfo{author}{H.~Fang}, \bibinfo{author}{C.~Yuan},
\newblock \bibinfo{title}{Convolution meets lora: Parameter efficient finetuning for segment anything model},
\newblock \bibinfo{journal}{arXiv preprint arXiv:2401.17868}  (\bibinfo{year}{2024}).
\bibitem[{Wei et~al.(2024)Wei, Chen, Jin, Ma, Liu, Ling, Wang, Chen, and Zheng}]{wei2024stronger}
\bibinfo{author}{Z.~Wei}, \bibinfo{author}{L.~Chen}, \bibinfo{author}{Y.~Jin}, \bibinfo{author}{X.~Ma}, \bibinfo{author}{T.~Liu}, \bibinfo{author}{P.~Ling}, \bibinfo{author}{B.~Wang}, \bibinfo{author}{H.~Chen}, \bibinfo{author}{J.~Zheng},
\newblock \bibinfo{title}{Stronger fewer \& superior: Harnessing vision foundation models for domain generalized semantic segmentation},
\newblock in: \bibinfo{booktitle}{Proceedings of the IEEE/CVF conference on computer vision and pattern recognition}, \bibinfo{year}{2024}, pp. \bibinfo{pages}{28619--28630}.
\bibitem[{Lin et~al.(2023)Lin, Xiang, Zhang, Yang, Yan, and Yu}]{lin2023samus}
\bibinfo{author}{X.~Lin}, \bibinfo{author}{Y.~Xiang}, \bibinfo{author}{L.~Zhang}, \bibinfo{author}{X.~Yang}, \bibinfo{author}{Z.~Yan}, \bibinfo{author}{L.~Yu},
\newblock \bibinfo{title}{Samus: Adapting segment anything model for clinically-friendly and generalizable ultrasound image segmentation},
\newblock \bibinfo{journal}{arXiv preprint arXiv:2309.06824} \bibinfo{volume}{4} (\bibinfo{year}{2023}).
\bibitem[{Lin et~al.(2024)Lin, Xiang, Wang, Cheng, Yan, and Yu}]{lin2024samct}
\bibinfo{author}{X.~Lin}, \bibinfo{author}{Y.~Xiang}, \bibinfo{author}{Z.~Wang}, \bibinfo{author}{K.-T. Cheng}, \bibinfo{author}{Z.~Yan}, \bibinfo{author}{L.~Yu},
\newblock \bibinfo{title}{Samct: Segment any ct allowing labor-free task-indicator prompts},
\newblock \bibinfo{journal}{IEEE Transactions on Medical Imaging}  (\bibinfo{year}{2024}).
\bibitem[{Liang et~al.(2025)Liang, Shi, Pu, Wu, Chen, Zhou, Xu, Chen, Chang, and Li}]{liang2025mambasam}
\bibinfo{author}{P.~Liang}, \bibinfo{author}{L.~Shi}, \bibinfo{author}{B.~Pu}, \bibinfo{author}{R.~Wu}, \bibinfo{author}{J.~Chen}, \bibinfo{author}{L.~Zhou}, \bibinfo{author}{L.~Xu}, \bibinfo{author}{Z.~Chen}, \bibinfo{author}{Q.~Chang}, \bibinfo{author}{Y.~Li},
\newblock \bibinfo{title}{Mambasam: A visual mamba-adapted sam framework for medical image segmentation},
\newblock \bibinfo{journal}{IEEE Journal of Biomedical and Health Informatics}  (\bibinfo{year}{2025}).
\bibitem[{Wang et~al.(2024)Wang, Ye, Cheng, Li, Chen, Cai, He, and Zhuang}]{wang2024sam}
\bibinfo{author}{G.~Wang}, \bibinfo{author}{J.~Ye}, \bibinfo{author}{J.~Cheng}, \bibinfo{author}{T.~Li}, \bibinfo{author}{Z.~Chen}, \bibinfo{author}{J.~Cai}, \bibinfo{author}{J.~He}, \bibinfo{author}{B.~Zhuang},
\newblock \bibinfo{title}{Sam-med3d-moe: Towards a non-forgetting segment anything model via mixture of experts for 3d medical image segmentation},
\newblock in: \bibinfo{booktitle}{International Conference on Medical Image Computing and Computer-Assisted Intervention}, \bibinfo{organization}{Springer}, \bibinfo{year}{2024}, pp. \bibinfo{pages}{552--561}.
\bibitem[{Zhang et~al.(2024)Zhang, Su, Xu, and Jia}]{zhang2024improving}
\bibinfo{author}{H.~Zhang}, \bibinfo{author}{Y.~Su}, \bibinfo{author}{X.~Xu}, \bibinfo{author}{K.~Jia},
\newblock \bibinfo{title}{Improving the generalization of segmentation foundation model under distribution shift via weakly supervised adaptation},
\newblock in: \bibinfo{booktitle}{Proceedings of the IEEE/CVF Conference on Computer Vision and Pattern Recognition}, \bibinfo{year}{2024}, pp. \bibinfo{pages}{23385--23395}.
\bibitem[{Chai et~al.(2024)Chai, Jain, Mo, Liu, Yang, Li, Tateyama, Lin, and Chen}]{chai2024novel}
\bibinfo{author}{S.~Chai}, \bibinfo{author}{R.~K. Jain}, \bibinfo{author}{S.~Mo}, \bibinfo{author}{J.~Liu}, \bibinfo{author}{Y.~Yang}, \bibinfo{author}{Y.~Li}, \bibinfo{author}{T.~Tateyama}, \bibinfo{author}{L.~Lin}, \bibinfo{author}{Y.-W. Chen},
\newblock \bibinfo{title}{A novel adaptive hypergraph neural network for enhancing medical image segmentation},
\newblock in: \bibinfo{booktitle}{International Conference on Medical Image Computing and Computer-Assisted Intervention}, \bibinfo{organization}{Springer}, \bibinfo{year}{2024}, pp. \bibinfo{pages}{23--33}.
\bibitem[{Ni et~al.(2024)Ni, Chen, Zhai, Tang, and Wang}]{ni2024context}
\bibinfo{author}{Z.~Ni}, \bibinfo{author}{X.~Chen}, \bibinfo{author}{Y.~Zhai}, \bibinfo{author}{Y.~Tang}, \bibinfo{author}{Y.~Wang},
\newblock \bibinfo{title}{Context-guided spatial feature reconstruction for efficient semantic segmentation},
\newblock in: \bibinfo{booktitle}{European Conference on Computer Vision}, \bibinfo{organization}{Springer}, \bibinfo{year}{2024}, pp. \bibinfo{pages}{239--255}.
\bibitem[{Rahman et~al.(2024)Rahman, Munir, and Marculescu}]{rahman2024emcad}
\bibinfo{author}{M.~M. Rahman}, \bibinfo{author}{M.~Munir}, \bibinfo{author}{R.~Marculescu},
\newblock \bibinfo{title}{Emcad: Efficient multi-scale convolutional attention decoding for medical image segmentation},
\newblock in: \bibinfo{booktitle}{Proceedings of the IEEE/CVF Conference on Computer Vision and Pattern Recognition}, \bibinfo{year}{2024}, pp. \bibinfo{pages}{11769--11779}.
\bibitem[{Shao et~al.(2025)Shao, Zeng, Hou, and Yang}]{shao2025mcanet}
\bibinfo{author}{H.~Shao}, \bibinfo{author}{Q.~Zeng}, \bibinfo{author}{Q.~Hou}, \bibinfo{author}{J.~Yang},
\newblock \bibinfo{title}{Mcanet: Medical image segmentation with multi-scale cross-axis attention},
\newblock \bibinfo{journal}{Machine Intelligence Research} \bibinfo{volume}{22} (\bibinfo{year}{2025}) \bibinfo{pages}{437--451}.
\bibitem[{Perera et~al.(2024)Perera, Erzurumlu, Gulati, and Yilmaz}]{perera2024mobileunetr}
\bibinfo{author}{S.~Perera}, \bibinfo{author}{Y.~Erzurumlu}, \bibinfo{author}{D.~Gulati}, \bibinfo{author}{A.~Yilmaz},
\newblock \bibinfo{title}{Mobileunetr: a lightweight end-to-end hybrid vision transformer for efficient medical image segmentation},
\newblock in: \bibinfo{booktitle}{European Conference on Computer Vision}, \bibinfo{organization}{Springer}, \bibinfo{year}{2024}, pp. \bibinfo{pages}{281--299}.
\bibitem[{Li et~al.(2025)Li, Li, Xing, Liao, Wang, Dong, Qin, and Yuan}]{li2025msdunet}
\bibinfo{author}{X.~Li}, \bibinfo{author}{L.~Li}, \bibinfo{author}{X.~Xing}, \bibinfo{author}{H.~Liao}, \bibinfo{author}{W.~Wang}, \bibinfo{author}{Q.~Dong}, \bibinfo{author}{X.~Qin}, \bibinfo{author}{C.~Yuan},
\newblock \bibinfo{title}{Msdunet: A model based on feature multi-scale and dual-input dynamic enhancement for skin lesion segmentation},
\newblock \bibinfo{journal}{IEEE Transactions on Medical Imaging}  (\bibinfo{year}{2025}).
\bibitem[{Liu et~al.(2024)Liu, Zhu, Liu, Yu, Chen, and Gao}]{liu2024rolling}
\bibinfo{author}{Y.~Liu}, \bibinfo{author}{H.~Zhu}, \bibinfo{author}{M.~Liu}, \bibinfo{author}{H.~Yu}, \bibinfo{author}{Z.~Chen}, \bibinfo{author}{J.~Gao},
\newblock \bibinfo{title}{Rolling-unet: Revitalizing mlp’s ability to efficiently extract long-distance dependencies for medical image segmentation},
\newblock in: \bibinfo{booktitle}{Proceedings of the AAAI conference on artificial intelligence}, volume~\bibinfo{volume}{38}, \bibinfo{year}{2024}, pp. \bibinfo{pages}{3819--3827}.
\bibitem[{Jiang et~al.(2025)Jiang, Zhang, Liu, Gao, Hu, Yan, Huang, and Liu}]{jiang2025rwkv}
\bibinfo{author}{J.~Jiang}, \bibinfo{author}{J.~Zhang}, \bibinfo{author}{W.~Liu}, \bibinfo{author}{M.~Gao}, \bibinfo{author}{X.~Hu}, \bibinfo{author}{X.~Yan}, \bibinfo{author}{F.~Huang}, \bibinfo{author}{Y.~Liu},
\newblock \bibinfo{title}{Rwkv-unet: Improving unet with long-range cooperation for effective medical image segmentation},
\newblock \bibinfo{journal}{arXiv preprint arXiv:2501.08458}  (\bibinfo{year}{2025}).
\bibitem[{Fu et~al.(2025)Fu, Lou, and Yu}]{fu2025segman}
\bibinfo{author}{Y.~Fu}, \bibinfo{author}{M.~Lou}, \bibinfo{author}{Y.~Yu},
\newblock \bibinfo{title}{Segman: Omni-scale context modeling with state space models and local attention for semantic segmentation},
\newblock in: \bibinfo{booktitle}{Proceedings of the Computer Vision and Pattern Recognition Conference}, \bibinfo{year}{2025}, pp. \bibinfo{pages}{19077--19087}.
\bibitem[{Li et~al.(2025)Li, Liu, Li, Wang, Liu, Liu, Chen, and Yuan}]{li2025u}
\bibinfo{author}{C.~Li}, \bibinfo{author}{X.~Liu}, \bibinfo{author}{W.~Li}, \bibinfo{author}{C.~Wang}, \bibinfo{author}{H.~Liu}, \bibinfo{author}{Y.~Liu}, \bibinfo{author}{Z.~Chen}, \bibinfo{author}{Y.~Yuan},
\newblock \bibinfo{title}{U-kan makes strong backbone for medical image segmentation and generation},
\newblock in: \bibinfo{booktitle}{Proceedings of the AAAI Conference on Artificial Intelligence}, volume~\bibinfo{volume}{39}, \bibinfo{year}{2025}, pp. \bibinfo{pages}{4652--4660}.
\bibitem[{Ruan et~al.(2024)Ruan, Li, and Xiang}]{ruan2024vm}
\bibinfo{author}{J.~Ruan}, \bibinfo{author}{J.~Li}, \bibinfo{author}{S.~Xiang},
\newblock \bibinfo{title}{Vm-unet: Vision mamba unet for medical image segmentation},
\newblock \bibinfo{journal}{ACM Transactions on Multimedia Computing, Communications and Applications}  (\bibinfo{year}{2024}).
\bibitem[{Ma et~al.(2024)Ma, He, Li, Han, You, and Wang}]{ma2024segment}
\bibinfo{author}{J.~Ma}, \bibinfo{author}{Y.~He}, \bibinfo{author}{F.~Li}, \bibinfo{author}{L.~Han}, \bibinfo{author}{C.~You}, \bibinfo{author}{B.~Wang},
\newblock \bibinfo{title}{Segment anything in medical images},
\newblock \bibinfo{journal}{Nature Communications} \bibinfo{volume}{15} (\bibinfo{year}{2024}) \bibinfo{pages}{654}.
\bibitem[{Codella et~al.(2019)Codella, Rotemberg, Tschandl, Celebi, Dusza, Gutman, Helba, Kalloo, Liopyris, Marchetti et~al.}]{codella2019skin}
\bibinfo{author}{N.~Codella}, \bibinfo{author}{V.~Rotemberg}, \bibinfo{author}{P.~Tschandl}, \bibinfo{author}{M.~E. Celebi}, \bibinfo{author}{S.~Dusza}, \bibinfo{author}{D.~Gutman}, \bibinfo{author}{B.~Helba}, \bibinfo{author}{A.~Kalloo}, \bibinfo{author}{K.~Liopyris}, \bibinfo{author}{M.~Marchetti}, et~al.,
\newblock \bibinfo{title}{Skin lesion analysis toward melanoma detection 2018: A challenge hosted by the international skin imaging collaboration (isic)},
\newblock \bibinfo{journal}{arXiv preprint arXiv:1902.03368}  (\bibinfo{year}{2019}).
\bibitem[{Ruan et~al.(2022)Ruan, Xiang, Xie, Liu, and Fu}]{ruan2022malunet}
\bibinfo{author}{J.~Ruan}, \bibinfo{author}{S.~Xiang}, \bibinfo{author}{M.~Xie}, \bibinfo{author}{T.~Liu}, \bibinfo{author}{Y.~Fu},
\newblock \bibinfo{title}{Malunet: A multi-attention and light-weight unet for skin lesion segmentation},
\newblock in: \bibinfo{booktitle}{2022 IEEE International Conference on Bioinformatics and Biomedicine (BIBM)}, \bibinfo{organization}{IEEE}, \bibinfo{year}{2022}, pp. \bibinfo{pages}{1150--1156}.
\bibitem[{Al-Dhabyani et~al.(2020)Al-Dhabyani, Gomaa, Khaled, and Fahmy}]{al2020dataset}
\bibinfo{author}{W.~Al-Dhabyani}, \bibinfo{author}{M.~Gomaa}, \bibinfo{author}{H.~Khaled}, \bibinfo{author}{A.~Fahmy},
\newblock \bibinfo{title}{Dataset of breast ultrasound images},
\newblock \bibinfo{journal}{Data in brief} \bibinfo{volume}{28} (\bibinfo{year}{2020}) \bibinfo{pages}{104863}.
\bibitem[{Sirinukunwattana et~al.(2017)Sirinukunwattana, Pluim, Chen, Qi, Heng, Guo, Wang, Matuszewski, Bruni, Sanchez et~al.}]{sirinukunwattana2017gland}
\bibinfo{author}{K.~Sirinukunwattana}, \bibinfo{author}{J.~P. Pluim}, \bibinfo{author}{H.~Chen}, \bibinfo{author}{X.~Qi}, \bibinfo{author}{P.-A. Heng}, \bibinfo{author}{Y.~B. Guo}, \bibinfo{author}{L.~Y. Wang}, \bibinfo{author}{B.~J. Matuszewski}, \bibinfo{author}{E.~Bruni}, \bibinfo{author}{U.~Sanchez}, et~al.,
\newblock \bibinfo{title}{Gland segmentation in colon histology images: The glas challenge contest},
\newblock \bibinfo{journal}{Medical image analysis} \bibinfo{volume}{35} (\bibinfo{year}{2017}) \bibinfo{pages}{489--502}.

\end{thebibliography}

\end{document}